\documentclass[letterpaper]{article} 
\usepackage{aaai25}  
\usepackage{times}  
\usepackage{helvet}  
\usepackage{courier}  
\usepackage[hyphens]{url}  
\usepackage{graphicx} 
\usepackage{natbib}  
\usepackage{caption} 
\usepackage{algorithm}
\usepackage{algorithmic}

\usepackage{newfloat}
\usepackage{listings}
\DeclareCaptionStyle{ruled}{labelfont=normalfont,labelsep=colon,strut=off} 
\floatstyle{ruled}
\newfloat{listing}{tb}{lst}{}
\floatname{listing}{Listing}
\title{Structured Packing in LLM Training Improves Long Context Utilization}
\author {
    Konrad Staniszewski\textsuperscript{\rm 1, \rm 2},
    Szymon Tworkowski\textsuperscript{\rm 1, \rm 6},
    Sebastian Jaszczur\textsuperscript{\rm 1, \rm 2},
    Yu Zhao\textsuperscript{\rm 3},
    Henryk Michalewski\textsuperscript{\rm 1, \rm 4},
    Łukasz Kuciński\textsuperscript{\rm 1, \rm 2, \rm 5},
    Piotr Miłoś\textsuperscript{\rm 1, \rm 2, \rm 5} 
}
\affiliations {
    \textsuperscript{\rm 1}University of Warsaw, Krakowskie Przedmiescie 26/28, 00-927 Warsaw, Poland\\
    \textsuperscript{\rm 2}IDEAS NCBR, Chmielna 69, 00-801 Warsaw, Poland\\
    \textsuperscript{\rm 3}University of Edinburg, Old College, South Bridge, Edinburgh EH8 9YL, United Kingdom\\
    \textsuperscript{\rm 4}Google DeepMind, 5 New Street Square, London, United Kingdom\\
    \textsuperscript{\rm 5}Institute of Mathematics Polish Academy of Sciences, Jana i Jedrzeja Sniadeckich 8, 00-656, Warsaw, Poland\\
    \textsuperscript{\rm 6}xAI, San Francisco Bay Area, California, U.S.\\
    ks.staniszewski@uw.edu.pl
}

\usepackage{multirow}
\usepackage{booktabs}
\usepackage{amstext}
\usepackage{subcaption}
\newcommand{\expnumber}[2]{{#1}\mathrm{e}{#2}}
\newcommand{\method}{\textsc{{SPLiCe}}}
\newcommand{\methodnorm}{{SPLiCe}}
\newcommand{\contrieverms}[1]{\textsc{Cont}}
\newcommand{\bmtf}[1]{\textsc{BM25}}
\newcommand{\repo}[1]{\textsc{Repo}}
\newcommand{\methodvariant}[1]{{\small\method \,{#1}}}
\newcommand{\us}[1]{{$_{(#1)}$}}

\newcommand{\apdx}[2]{#2}

\begin{document}

\maketitle

\begin{abstract}
    Recent advancements in long-context language modeling have attracted significant attention, yet their practical applications often suffer from suboptimal context utilization. To efficiently address this issue, we introduce the Structured Packing for Long Context, \method{}, a method that uses retrieval to collate mutually relevant documents into long training samples. 
    We demonstrate that \method{} improves performance on long-context tasks, particularly by achieving perfect accuracy on the synthetic Needle in the Haystack benchmark, and effectively mitigating the ‘lost-in-the-middle’ phenomenon often observed in large language models. Notably, these long-context capabilities also extend to realistic downstream tasks, such as Qasper, across multiple model sizes—3B, 7B, and 13B—and are achieved with only brief fine-tuning on 2-6 billion tokens.
    We supplement these results with a detailed analysis of \method{}, examining the impact of hyperparameter choices, the different mixtures and proportions of \method{}-generated training data, and the choice of the retriever. We also study the transfer of long-context utilization skills between the modalities.
    An intriguing finding from our analysis is that training on a corpus of code can enhance performance on natural language tasks.
\end{abstract}

%
\begin{links}
    \link{Code}{https://github.com/ideas-ncbr/publications_2024}
    \link{Extended version}{https://arxiv.org/abs/2312.17296}
\end{links}

\section{Introduction}
Large language models (LLMs) \citep{brown2020language, chowdhery2022palm, lewkowycz2022solving, openai2023gpt4, bai2023qwen} have transformed the field of AI. Recently, the field has observed the rise of Long Context Language Models (LCLMs) that promise to unveil novel and powerful capabilities \citep{anthropic2023modelcard, gpt4-turbo, reid2024gemini}. However, their ability to process long contexts is not always as effective as one hopes. Indeed, several studies have highlighted an important limitation: when processing prompts composed of multiple documents, LCLMs frequently encounter difficulties in accurately extracting relevant information~\citep{tworkowski2023focused, liu2023lost, Shi2023LargeLM,niths}. Additionally, they typically find it challenging to utilize information from the middle of their inputs~\citep{liu2023lost}, even on simple synthetic retrieval tasks~\citep{longchat2023}. Understanding these issues is vital for advancements in LCLM technologies and calls for systematic research.  

In this work, we take a step towards better context utilization in LCLMs. We focus on training data, keeping other components, such as the architecture and training objectives, unchanged. The broad question is: \emph{Given training data consisting of documents, how should these documents be organized into training samples to enhance long-context capabilities}? While this perspective has received some attention recently \citep{levine2022icl,data-distributional-properties,shi2023incontext}, the problem remains unsolved. The central finding of this work is that \emph{structuring training data to increase semantic interdependence is an effective strategy towards better long context utilization}. We achieve this by introducing and evaluating Structured Packing for Long Context (\method{}), a method for creating training samples by using retrieval (e.g., BM25, Contriever) to collate mutually relevant documents into a single training context.
    
\begin{figure}
    \centering
    \includegraphics[width=227.65pt]{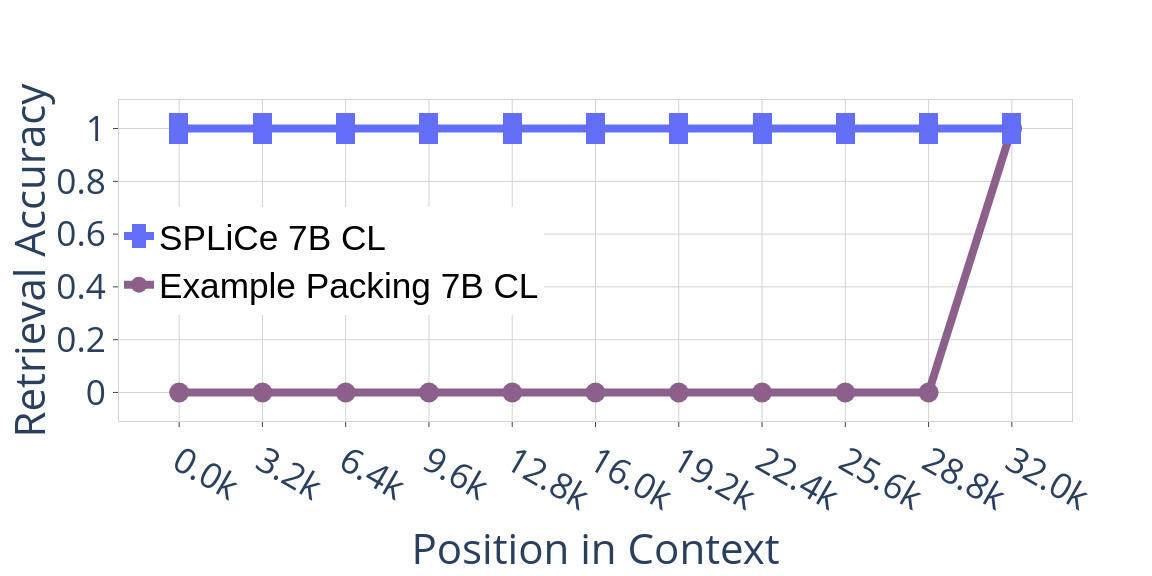}
    \caption{\method{} vs \textsc{Example Packing (EP) (baseline)} on Needle in a Haystack. A model fine-tuned with \method{} achieves perfect accuracy in retrieving fine-grained information over the whole context, while the baseline can only handle a small final segment (details in Appendix \apdx{sec:appendix_niths}{N}).}
    \label{fig:splice_vs_baseline_niths}
\end{figure}

We empirically validate \method{} showing that \mbox{fine-tuning} of OpenLLaMA $3$Bv2, $7$Bv2 \citep{openlm2023openllama} and CodeLlama $13$B  \citep{roziere2023code} with mere $2$B--$6$B tokens already brings improvements in handling long context information in downstream tasks that require retrieval and in-context learning. {These tasks include Qasper \citep{dasigi2021dataset} from SCROLLS \citep{shaham2022scrolls}}, HotPotQA \citep{hotpotqa}, Needle In A Haystack \citep{niths}, TREC \citep{li-roth-2002-learning, hovy-etal-2001-toward}, and DBpedia \citep{dbpedia}. We also show that \method{} significantly alleviates the 'lost-in-the-middle' phenomenon \citep{liu2023lost} and outperforms standard example packing on the Needle In A Haystack task \citep{niths} {(see Figure \ref{fig:splice_vs_baseline_niths})}.  We perform a comprehensive study of the design choices and properties of \method{}, showing, in particular, that the acquired long context capabilities transfer between modalities, such as code and text. \method{} also helps to retain and in some cases even improve performance on short context benchmarks like GSM8K \citep{cobbe2021training}, MMLU \citep{hendrycks2021measuring} and TREC \citep{li-roth-2002-learning, hovy-etal-2001-toward}.

Our contributions can be summarized as follows:
\begin{itemize}
    \item We comprehensively show that structuring training data is a viable way of improving the long context utilization. To this end, we introduce \method{}, a method for creating training samples by using retrieval to collate mutually relevant documents into a single sample.
    \item We fine-tune OpenLLaMA $3$Bv2, OpenLLaMA $7$Bv2~\citep{openlm2023openllama} and CodeLlama $13$B  \citep{roziere2023code} using \method{}, showing that it improves long-contex downstream performance.
    \item We provide a comprehensive analysis of \method{}’s design choices, including retrieval parameters and document concatenation order, and evaluate its robustness and scalability with varying data sources and a parametrizable noisy retriever.
\end{itemize}

\section{Method}\label{sec:methods}

\method{} is a method for constructing training samples that improve the effectiveness of long-context fine-tuning.
This leads to improved performance in tasks such as in-context learning, question answering, information retrieval, and long-context language modeling ({see Section \ref{sec:large_scale}}).

\paragraph*{Rationale and Intuitions}{
Capturing long-range dependencies is believed to enhance language modeling and retrieval-augmentation \citep{borgeaud2022retro}. It is an open question how to achieve such benefits in pre-training or fine-tuning. The primary difficulty comes from long-range dependencies being rare in training data \citep{deVries2023} and diminishing with distance. Thus, it is unlikely that a model will learn to utilize long context without more guidance.

Recent studies indicate that structuring data, i.e., going beyond the i.i.d. paradigm, might be beneficial or even necessary to achieve good long-context performance. \citep{levine2022icl} develops a theory showing that the trained model establishes stronger dependencies between text segments in the same training sample. 
Whereas concurrently to our work \citep{shi2023incontext} shows that pre-training on structured data improves performance (see also Section \ref{sec:related_work} for a more detailed comparison).
\method{} follows these intuitions, and  constructs training samples by concatenating mutually relevant documents to increase the dependency density, thus allowing the model to learn to utilize long context.}

\paragraph*{Structured Packing for Long Context (\method)}{
\method{} starts by picking a random document from the dataset to create a {root of a tree} and continues in a breadth-first manner, each time appending top-$k$ similar documents from the corpus. The final sequence is generated by flattening the tree according to a specific traversal strategy; see Algorithm \ref{alg:splice}. 
The hyperparameter $k$ introduces flexibility, enabling interpolation between different retrieval modes. Specifically, when $k=1$, \method{} simulates a long document by creating a path of related examples. For larger $k$ values, \method{} generates examples akin to those used in retrieval-augmented models, e.g. \citep{borgeaud2022retro}.
}

\paragraph*{\method{} Retrieval}{
Many possible retrieval methods can be used with \method{} (\textsc{Retrieve} function in Algorithm \ref{alg:splice}). In our experiments, we test the following:
\begin{itemize}
  \item \methodvariant{\repo{}}: based on additional meta-information about the data, that is the repository structure of the code (\repo{}): we concatenate files using a depth-first search algorithm on the directory structure, that is files from the same directory are grouped together. A similar method has been pioneered by \citep{wu2022memorizing} and proposed in \citep{shi2023incontext} as an interesting future direction.
  
  \item \methodvariant{\bmtf{}}: based on \bmtf{} \citep{robertson2009probabilistic, Bassani_retriv_A_Python_2023}, a standard retrieval method that uses a bag-of-words approach to rank documents based on their similarity to a query. 
  
  \item \methodvariant{\contrieverms{}}: based on {Contriever-MSMARCO} (\contrieverms{}) \citep{izacard2022unsupervised}, a retrieval method that uses a transformer to rank documents based on their similarity.
\end{itemize}
}

\begin{figure}
    \centering
    \includegraphics[width=227.65pt]{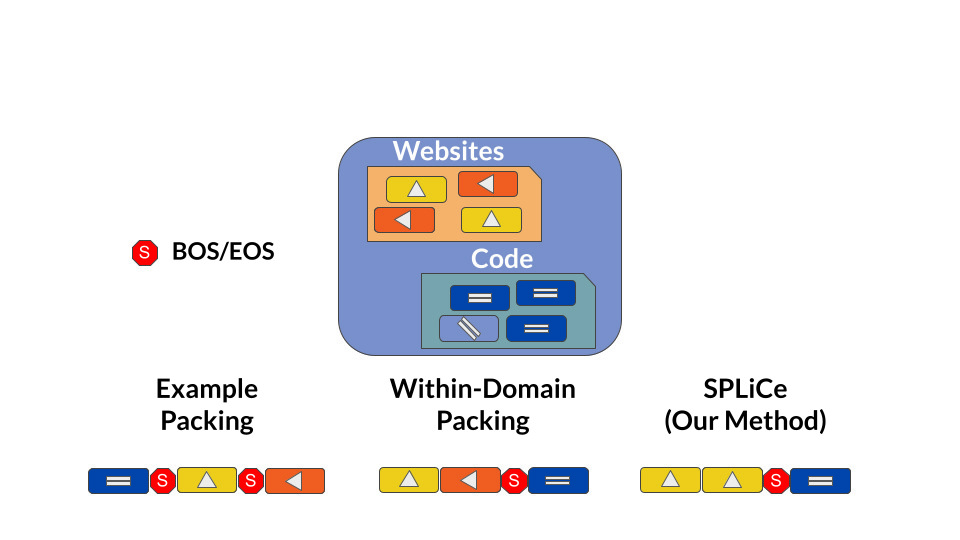}
    \caption{Training samples generated by Example Packing, Within-Domain Example packing, and \method{}. Similar colors and shapes indicate related documents, which could be found using a retrieval method (e.g., BM25 or Contriever) or metadata (e.g., git repository structure).}
    \label{fig:standard_vs_splice}
\end{figure}

\begin{figure}
    \begin{center}
      \includegraphics[scale=.28]{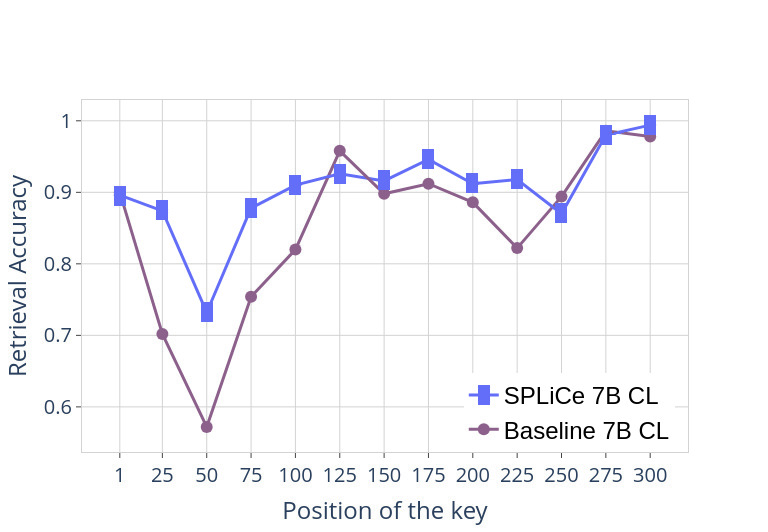}
    \end{center}
    \caption{Key-value retrieval performance on a dictionary of $300$ key-value pairs ($\approx$ $24$K tokens). The $7$B CL model trained with \method{} achieves much higher accuracy on hard-to-retrieve positions in the middle than the Example Packing Baseline. The details about this task can be found in Appendix \apdx{sec:appendix_kv_retriv_nelson}{D}.
    Each position averaged over $500$ examples.}
    \label{fig:lit300k7bcl}
\end{figure}

\begin{algorithm}
    \caption{\method{} training sample construction } \label{alg:splice}
    \begin{algorithmic}
    \STATE \textbf{Input:} \\
        $\quad$$D$: document corpus\\
        $\quad$$k$: breadth hyper-parameter\\
        $\quad$$L$: maximum length of returned training sample\\
        $\quad$$\texttt{RETRIEVE}$: retrieval method to use, e.g., BM25\\
        $\quad$$\texttt{ORDER}$: ordering method, e.g., identity, or shuffle\\
    \STATE \textbf{Output:} 
        training sample
    \STATE $d_r \sim D$ \COMMENT{Sample the root document}
    \STATE $D = D \setminus \{d_r\}$ 
    \STATE $C = [d_r]$ 
    \STATE $Q = \texttt{empty queue }$ 
    \STATE $Q.\texttt{PUSH}(d_r)$
    \WHILE{$Q\neq \emptyset$ and $\texttt{len}(C) \leq L$}
      \STATE $d = Q.\texttt{POP}()$ 
      \STATE  $d_{1}, \ldots, d_{k} = \texttt{RETRIEVE}(d, k)$ 
      \STATE \COMMENT{Retrieve top-$k$ most similar documents to $d$ using a selected method, e.g., BM25}
        \FOR{each $d_{i}$ in $d_{1}, \ldots, d_{k}$}
          \IF{$d_{i} \in D$}  
            \STATE \COMMENT{$\texttt{RETRIEVE}$ uses a precomputed index and may return documents that are already in $C$}
            \STATE $C = C.\texttt{APPEND}(d_{i})$ \COMMENT{Append $d_{i}$ to $C$}
            \STATE  $Q.\texttt{PUSH}(d_{i})$ 
            \STATE $D = D \setminus \{d_i\}$
          \ENDIF
        \ENDFOR
    \ENDWHILE
    \STATE \textbf{return} $\texttt{CONCAT}(\texttt{TRIM}(\texttt{ORDER}(C), L))$ 
    \end{algorithmic}
\end{algorithm}

\paragraph*{\method{} Computational Efficiency}{ Given the dataset sizes used in training LLMs, computational efficiency plays a crucial role. \methodvariant{\repo{}} is the fastest and easiest to implement but requires additional directory structure, i.e., it does not apply to general web data. \methodvariant{\bmtf{}} uses a bag of words \bmtf{} method that lacks deeper semantic encoding. However, it was observed to have strong generalization properties \citep{thakur2021beirheterogenousbenchmarkzeroshot}. \methodvariant{\contrieverms{}} requires calculating embeddings for each document and retrieval based on the vector inner-product, but can have poorer generalization properties than \bmtf{} \citep{thakur2021beirheterogenousbenchmarkzeroshot}. The retrieval step can be done efficiently using a fast approximate max IP search, e.g., Faiss \citep{johnson2017billionscale}.
To reduce the number of times the training sample requires just copy-paste abilities and improve training step efficiency, we employ the StarCoder \citep{li2023starcoder} dataset, which was deduplicated using the pipeline from \citep{allal2023santacoder}.}

\section{Experiments} \label{sec:large_scale}

In this section, we show that \method{} improves the long context performance of large-scale language models. To this end, we use $3$B, $7$B, and $13$B parameter models. First, in Section \ref{sec:exps_big_scale}, we focus on tasks that test in-context learning, question answering, and in-context information retrieval. Next, we show that \method{} can improve the core model capabilities by testing its short context performance. Finally, in Section \ref{sec:exps_medium_scale}, we train over 40 medium-size models ($270$M parameters) using different data mixtures and \method{} parameters to analyze various design choices, robustness to noise, and scaling properties. 

An important finding of our work is that presented improvements occur during a relatively short fine-tuning. To be more precise, $3$B models were tuned on $5.4$B tokens, whereas $7$B and $13$B models were tuned on $2$B tokens.
\subsection{Baselines}
We consider two popular baselines used in LLM training pipelines.
The first one is Example Packing \citep{brown2020language}, used in the training of GPT-3 models.
It constructs training samples by randomly sampling documents from the dataset and separating them with BOS/EOS tokens.
The second one, which we call Within-Domain Example Packing takes random documents from the same meta-class (for example, Wikipedia, C source code) and concatenates them to create a training sample \citep{groeneveld2024olmo,zhao2024analysing}.
We compare \method{} against both baselines. As we note no clear benefit of Within-Domain Example Packing over Example Packing in fine-tuning case (see Table \apdx{tab:bigresultsrandbasreposplice}{21} in Appendix \apdx{sec:detailed_results}{B.3} ) in the main body of the paper we compare only against a more established Example Packing. We visualize the differences between baselines in Figure \ref{fig:standard_vs_splice}.

\subsection{Experimental Setup}\label{sec:exp_res_big}
For $3$B model experiments, we fine-tune on a 50/50 mixture of RedPajama, prepared in the standard way, and C prepared using \methodvariant{\bmtf{}}. For $7$B and $13$B ones, we fine-tune on a 50/25/25 mixture of RedPajama (50) prepared in the standard way, StackExchange (25) and C (25) prepared using \methodvariant{\bmtf{}}. StackExchange is part of RedPajama \citep{together2023redpajama}, and C data come from StarCoder \citep{li2023starcoder}. Including the standard RedPajama aims to prevent the model from overfitting to artificially created documents {and is inspired by \citep{ouyang2022training,roziere2023code}}. {We analyze the impact of data mixture in Section \ref{sec:exps_medium_scale}}.

We fine-tune with $32$K context length. We employ the Focused Transformer (FoT) \citep{tworkowski2023focused} and CodeLlama (CL) context extension methods \citep{roziere2023code}. We use a batch size of $256$K ($512$K, resp.) tokens per step for $3$B and $7$B  ($13$B, resp.) models. We set the learning rate of $\expnumber{1.5}{-5}$ with linear warmup and cosine decay, following \citep{openlm2023openllama}.
In the next section, we test \textbf{eight models}:
 \begin{center}
  {$\{$$3$B FoT, $7$B FoT, $7$B CL, $13$B FoT $\}$ $\times$ $\{$\method{}, \textsc{EP}$\}$,} 
\end{center}
 where \textsc{EP} denotes the standard Example Packing \citep{brown2020language} method (serving as baseline) where context is created by sampling random documents from the corpus and separating them with \texttt{BOS}/\texttt{EOS} tokens (see Figure \ref{fig:standard_vs_splice}). We provide results regarding the Within-Domain Example Packing baseline in Appendix \apdx{sec:appendix_boseos}{C.2}. If not stated otherwise, in \method{} we use $k=1$ and the identity permutation as \texttt{Order} in the Algorithm \ref{alg:splice}. Hyperparameter details can be found in Appendix \apdx{sec:appendix_architecture}{A}.

 \subsection{Experimental Results} \label{sec:exps_big_scale}

 \subsubsection{In-Context Learning} \label{sec:in_context_learning}

 In this section, we ask the following research questions: 
 \emph{Does \methodnorm{} improve in-context learning abilities? 
 If so, with what context length is it the case?} 
 To answer those questions, we evaluate the accuracy of our models on TREC \citep{li-roth-2002-learning, hovy-etal-2001-toward} and DBpedia~\citep{dbpedia}, which are text classification tasks. 
 For TREC we test $\{2, 16, 32\}$K context lengths, which correspond to $\{90, 780, 1560\}$ in context examples, respectively. 
 For DBpedia, we test $\{16, 32\}$K context lengths, which correspond to $\{190, 380\}$ in-context examples, respectively, and omit the 2K length due to its limited capacity of 20 in-context examples. 
 For each context length, we average the results across several sets of in-context examples and provide average improvement of \method{} and its $95$\% bootstrap confidence interval (improvements are calculated per set of in-context examples, see Appendix \apdx{sec:appendix_trec}{H}). 
 \textbf{In both tasks and all considered context lengths, we note that \method{} significantly improves in-context learning abilities in comparison to both Example Packing and the starting checkpoint.} We hypothesize that by increasing the amount of potentially relevant information in context, \method{} allows the model to learn longer and better context lookups. We further study this in Section \ref{sec:splice_data_properties}, where we analyze \method{} using the framework from \citep{data-distributional-properties}.
 We present the main results in Tables \ref{tab:table_trec}, \ref{tab:table_db} and additional in Table \apdx{tab:table_trec_db_short}{35} in Appendix \apdx{sec:appedix_trec_2_k}{O}. 
 Additionally in Appendix \apdx{sec:appendix_trec}{H} we analyze results per-set of in-context examples and show that \method{} achieves stochastic domination over Example Packing.

 \begin{table}[ht]

    \begin{center}
      \begin{tabular}{c c|ccc }
        \toprule
         \multicolumn{5}{c}{\textbf{TREC}} \\
        \midrule
         Model & Context  &  \textsc{EP} &  \method{} & $\Delta ${\scriptsize~[conf interv]} \\
        \midrule
        
        \multirow{2}{*}{$3$B$_{ \text{FoT}}$} 
        & $32$K & 73.9  & \textbf{79.3} & 5.4~\scriptsize[${4.7,6.2}]$~~ \\
        & $16$K & 68.9  & \textbf{76.9} & 8.0~\scriptsize[${6.9,9.3}]$~~ \\
 
        \midrule    
        
        \multirow{2}{*}{$7$B$_{ \text{FoT}}$} 
        & $32$K & 75.6  & \textbf{79.4} & 3.8~\scriptsize[${2.1,5.1}]$ \\
        & $16$K& 74.0  & \textbf{79.0} & 5.0~\scriptsize[${3.4,6.0}]$ \\
 
        \midrule
        
        \multirow{2}{*}{$7$B$_{ \text{CL}}$} 
        & $32$K & 75.3  & \textbf{76.6} & 1.3~\scriptsize[${0.8,1.8}]$ \\
        & $16$K & 81.4  & \textbf{82.5} & 1.1~\scriptsize[${0.2,1.6}]$ \\
 
        \midrule
        
        \multirow{2}{*}{$13$B$_{ \text{FoT}}$} 
        & $32$K & 89.2  & \textbf{92.4} & 3.2~\scriptsize[${2.6,3.8}]$\\
        & $16$K & 88.2  & \textbf{91.2}	 & 3.0~\scriptsize[${1.9,3.5}]$ \\
 
        \bottomrule
        \end{tabular}
    \end{center}
    \caption{ We test the classification accuracy on TREC \citep{li-roth-2002-learning, hovy-etal-2001-toward}. We average across $50$ sets of in-context examples for $3$B  models, $10$ for $7$B models, and $5$ for $13$B models. $\Delta${\scriptsize~[ci]} denotes the mean improvement and its $95\%$ bootstrap confidence intervals (see Appendix \apdx{sec:appendix_trec}{H}).}\label{tab:table_trec}
  \end{table}%

  \begin{table}[ht]

    \begin{center}
      \begin{tabular}{c c|ccc }
        \toprule

         \multicolumn{5}{c}{\textbf{DBpedia}} \\
        \midrule
         Model & Context  &  \textsc{EP} &  \method{} & $\Delta ${\scriptsize~[conf interv]}\\
        \midrule
        
        \multirow{2}{*}{$3$B$_{ \text{FoT}}$} 
        & $32$K & 82.9  & \textbf{85.9} & 3.0~\scriptsize[${2.6,3.6}]$ \\
        & $16$K & 79.1  & \textbf{82.0} & 2.9~\scriptsize[${2.5,3.4}]$\\

        \midrule    
        
        \multirow{2}{*}{$7$B$_{ \text{FoT}}$} 
        & $32$K & 82.9  & \textbf{84.9} & 2.0~\scriptsize[${1.5,2.4}]$ \\
        & $16$K & 83.6  & \textbf{85.6} & 2.0~\scriptsize[${1.5,2.5}]$ \\
        
        \midrule
        
        \multirow{2}{*}{$7$B$_{ \text{CL}}$} 
        & $32$K & 95.1  & \textbf{95.6} & 0.5~\scriptsize[${0.3,0.6}]$ \\
        & $16$K & 96.2  & \textbf{96.4} & 0.2~\scriptsize[${0.0,0.3}]$ \\
        \midrule
        
        \multirow{2}{*}{$13$B$_{ \text{FoT}}$} 
        & $32$K & 95.6 & \textbf{96.0} & 0.4~\scriptsize[${0.0,0.8}]$ \\
        & $16$K & 95.8 & \textbf{96.8} & 1.0~\scriptsize[${0.6,1.5}]$ \\

        \bottomrule
        \end{tabular}
    \end{center}
    \caption{ We average results across $40$ sets of in-context examples for  $3$B and $7$B models and $5$ for $13$B. Due to the size of the DBpedia dataset, for each set of in-context examples, we sample a subset of $500$ elements of the evaluation set.}\label{tab:table_db}
  \end{table}%

  \subsubsection{Question Answering and In-Context Retrieval} \label{sec:question_answering}

  In this section, we ask the following research question: \emph{Does fine-tuning on \methodnorm{} prepared data result in improved question-answering abilities?}
  To answer the question, we utilize popular long context benchmarks such as Needle In A Haystack \citep{niths} and lost-in-the-middle key-value retrieval \citep{liu2023lost}, along with Qasper \citep{dasigi2021dataset} from SCROLLS \citep{shaham2022scrolls}, HotPotQA \citep{hotpotqa} passkey \citep{mohtashami2023landmark} and parts of RULER \citep{hsieh2024rulerwhatsrealcontext} tasks.

  On Needle In A Haystack, we observe that the model fine-tuned on data prepared by \method{} strongly outperforms the model fine-tuned on data prepared by Example Packing. To be more precise \textbf{model fine-tuned with \method{} can answer the question no matter the location of the relevant piece of information.}
  Whereas the model trained with Example Packing only manages to answer correctly when the information is close to the question (see Figure \ref{fig:splice_vs_baseline_niths}).
   We also test our models on the lost-in-the-middle key-value retrieval task  \citep{liu2023lost}, and observe that \method{} helps on hard-to-retrieve positions (see Figure \ref{fig:lit300k7bcl}). The main difference between those two tasks is that in the lost-in-the-middle key-value retrieval task, the input is highly structured (dictionary of random $128$ bit UUIDs, see Appendix~\apdx{sec:appendix_kv_retriv_nelson}{D} for details) and the objective of the model is to retrieve the value assigned to a given key. On the other hand, in the Needle In A Haystack, a piece of information is placed inside a large coherent text, and the model is asked a question related to this information (see Appendix \apdx{sec:appendix_niths}{N} for details).
  
  We additionally evaluate our models on Qasper \citep{dasigi2021dataset}, HotPotQA \citep{hotpotqa} passkey retrieval \citep{mohtashami2023landmark} and RULER \citep{hsieh2024rulerwhatsrealcontext} and observe that \method{} results in improvements over both the Example Packing and starting checkpoint. We present the results in Appendix \apdx{sec:appendix_qasper_hpqa}{K}.

\subsubsection{Short Context Evaluation} \label{sec:short_context}
One challenge in long-context fine-tuning is the degradation of short-context performance \citep{dubey2024llama3herdmodels}. This can be overcomed by upsampling the short-context data and more gradual context extension \citep{dubey2024llama3herdmodels}. We note that those approaches are compatible with \method{} and instead focus on comparing \method{} with Example Packing in a single-step context extension setup. We observe that \method{} seems to be either better or on par with Example Packing (see Table \ref{tab:table_shortctx}). What is intriguing is that for the $13$B parameter model, \method{} even improves the short context performance on GSM8K \citep{cobbe2021training} by ($+1.7$) over the starting checkpoint. We hypothesize that GSM8K is a much more attention-demanding task than MMLU \citep{hendrycks2021measuring}, as it requires extracting relevant pieces of information, composing a chain of thought, and writing the final answer. Whereas MMLU is a well-established collection of tests spanning multiple domains. We hypothesize that the improvement does not occur in smaller models due to their low scores on GSM8K, as we get similar results when evaluating on code in Appendix \apdx{sec:apendix_human_eval}{L}.

\begin{table}
    \centering

    \begin{tabular}{l|ccc|c}
    \toprule
    \multicolumn{1}{c}{\textbf{Model}}& \multicolumn{3}{c|}{\textbf{MMLU}} & \textbf{GSM8K}\\
    & STEM & HUM & \multicolumn{1}{c|}{All} &  \\
    \midrule
    $7$B \textsc{ST CHKP} & \textbf{33.6} & \textbf{42.1} & \textbf{40.8}  & \textbf{8.0} \\
    $7$B \textsc{2K Tuned}  & \textbf{33.7} & 40.8  & 39.4 & \textbf{8.4}  \\
    $7$B$_{\text{FoT}}$ \method{} & 31.0 & 35.6 & 36.3 & \textbf{7.6}  \\
    $7$B$_{\text{FoT}}$ \textsc{EP} & 30.1 & 36.8 & 36.2 & 6.7  \\
    $7$B$_{\text{CL}}$ \method{} & \textbf{32.7} & 38.8  & 36.5 & 5.9  \\
    $7$B$_{\text{CL}}$ \textsc{EP} & \textbf{32.7} & 37.5 & 36.1 & 6.3 \\
    \midrule
    $13$B \textsc{ST CHKP} & 36.6 & \textbf{48.2}  & \textbf{44.3} & 21.4  \\
    $13$B$_{\text{FoT}}$ \method{} & \textbf{38.3} & \textbf{48.6}  & \textbf{44.4} & \textbf{23.1} \\
    $13$B$_{\text{FoT}}$ \textsc{EP} & \textbf{39.0} & 47.5  & \textbf{44.7} & 21.9  \\
    \bottomrule
    \end{tabular}
    \caption{ We evaluate our models on MMLU ($5$-shot), GSM8K ($8$-shot CoT). We provide an additional comparison with their starting checkpoint. For the $7$B case, we additionally compare with a model tuned with $2$k context length on the same data. For each task, we highlight the best results up to $1$ point. For $3$B model results see Appendix \apdx{sec:apendix_table_shortctx_3B}{I}.}\label{tab:table_shortctx}
\end{table}

\subsection{Detailed Study with Medium Models} \label{sec:exps_medium_scale}
In Table \ref{tab:secondResults}, Table \ref{tab:cvariancecheck}, and Table \ref{tab:data_proportions}, we present a comprehensive examination of the impact of document packing on long-context performance using $270$M parameter models, showing that \method{} brings consistent improvement in long context language modeling. In Table \ref{tab:csharpsek}, we scale context to 64K and observe even greater benefits over the Example Packing. We also expand our results to 131K and 160K context length in Tables \apdx{tab:csharpsek128}{19} and \apdx{tab:csharpsek128160}{20} in Appendix. In Table \ref{tab:data_noisy}, we show that \method{} is quite robust to the non-accurate retriever. {What is more results in Table \ref{tab:data_noisy} clearly show that \method{} is an improvement over Within-Domain Example Packing. In particular, we note that the more noise we add, the closer \method{} is to Within-Domain Example Packing (semantically), and that with 100\% noise \method{} turns into Within-Domain Example Packing (this is because we use \method{} to prepare data coming from a single domain).} 

\paragraph*{Training and Evaluation}
Initially, we train with the $2$K context length on $6.3$B tokens from RedPajama \citep{together2023redpajama}. Subsequently, we fine-tune using $1$B tokens with the context extended to $32$K on a mixture of the original RedPajama data \citep{together2023redpajama} and long context data created using \method{}/\textsc{EP}. We employ the Focused Transformer (FoT) \citep{tworkowski2023focused} for context extension (unless stated otherwise). We measure perplexity on held-out portions of the arXiv \citep{azerbayev2022proofnet} and StarCoder \citep{li2023starcoder} datasets. The selection of these datasets is motivated by the fact that they can benefit from long-context information as demonstrated in \citep{chen2023extending, li2023starcoder}. 
We provide additional details in Appendix \apdx{sec:appendix_training_eval_proto}{M}.

\begin{table*}

    \begin{center}
    \begin{tabular}{c c | c | c c c c | c}
      \toprule
      \textbf{Altered} & \textbf{Method} & \multicolumn{1}{c}{\textbf{arXiv}} & \multicolumn{4}{c}{\textbf{Code}} & \multicolumn{1}{c}{\textbf{Code \&}} \\
      \textbf{Data} &  &  & Haskell  & Python & CUDA & All & \textbf{arXiv}  \\
  
      \midrule
  
      \multirow{3}{*}{C\#} 
      & \methodvariant{\bmtf{}} & \bf{5.52} \us{ .13} & \bf{3.33} \us{ .25} & \bf{2.90} \us{ .17} & \bf{2.46} \us{ .19} & \bf{3.11} \us{ .20} & \bf{3.26} \us{ .20}\\
      & \methodvariant{\contrieverms{}} & 5.53\us{.12} & 3.35\us{.23} & 2.91\us{.16} & 2.48\us{.17} & 3.12\us{.19} & 3.27\us{.19} \\
      & \methodvariant{\repo{}} & 5.53 \us{ .12} & 3.35 \us{ .23} & 2.91 \us{ .16} & 2.49 \us{ .16} & 3.12 \us{ .19} & 3.27 \us{ .19} \\
      & \textsc{EP}  & 5.65 & 3.58 & 3.07 & 2.65 & 3.31 & 3.46 \\
      
      \midrule
      \multirow{3}{*}{Python} 
      & \methodvariant{\bmtf{}} & \textbf{5.47} \us{ .10} & \textbf{3.25} \us{ .21} & \textbf{2.53} \us{ .09} & \textbf{2.41} \us{ .15} & \textbf{3.02} \us{ .15} & \textbf{3.17} \us{ .15}\\
      & \methodvariant{\contrieverms{}} & 5.49 \us{ .08} & 3.28 \us{ .18} & \textbf{2.53} \us{ .09} & 2.43 \us{ .13} & 3.03 \us{ .14} & 3.19 \us{ .13} \\
      & \methodvariant{\repo{}} & 5.48 \us{ .09} & 3.27 \us{ .19} & 2.54 \us{ .08} & 2.44 \us{ .12} & 3.03 \us{ .14} & 3.18 \us{ .14} \\
      & \textsc{EP} & 5.57 & 3.46 & 2.62 & 2.56 & 3.17 & 3.32 \\
      
      \midrule
      \multirow{3}{*}{{Wikipedia}} 
      & \methodvariant{\bmtf{}} & \textbf{5.64} \us{ .09} & \textbf{3.82} \us{ .15} & \textbf{3.26} \us{ .11} & \textbf{2.87} \us{ .13} & \textbf{3.55} \us{ .13} & \textbf{3.68} \us{ .13} \\
      & \methodvariant{\contrieverms{}} & 5.65 \us{ .08} & 3.87 \us{ .10} & 3.30 \us{ .07} & 2.92 \us{ .08} & 3.59 \us{ .09} & 3.72 \us{ .09} \\
      & \textsc{EP} & 5.73 & 3.97 & 3.37 & 3.00 & 3.68 & 3.81 \\
      \midrule
      \multirow{3}{*}{StackEX} 
      & \methodvariant{\bmtf{}} & \textbf{5.07} \us{ .07} & \textbf{3.88} \us{ .06} & \textbf{3.32} \us{ .04} & \textbf{2.89} \us{ .05} & \textbf{3.60} \us{ .05} & \textbf{3.69} \us{ .05} \\
       & \methodvariant{\contrieverms{}} & 5.09 \us{ .05} & 3.91 \us{ .03} & 3.35 \us{ .01} & 2.93 \us{ .01} & 3.63 \us{ .02} & 3.73 \us{ .01}\\
      & \textsc{EP} & 5.14 & 3.94 & 3.36 & 2.94 & 3.65 & 3.74 \\
  
      \bottomrule
    \end{tabular}
    \end{center}
    \caption{ Perplexity with an improvement over \textsc{EP} highlighted in the subscript:$_{(\text{imp over \textsc{EP}})}$. We fine-tune a $270$M parameter model with $32$K context on a 50/50 mixture of RedPajama 
    (organized in a standard way) and long-context data C\#, Python, Wikipedia, StackExchange prepared using a method of choice (\methodvariant{\bmtf{}}, \methodvariant{\contrieverms{}}, \methodvariant{\repo{}}, \textsc{EP}). \textsc{EP} denotes organizing long-context data in the same way as RedPajama.
    \method{} beats the \textsc{EP} often by a large margin. The variants of \method{} perform similarly, with \methodvariant{\bmtf{}} being slightly better.
    For detailed results, see Appendix \apdx{sec:detailed_results}{B.3}. }\label{tab:secondResults}

    \begin{center}

    \begin{tabular}{c c | c | c c c c | c}
      \toprule
      \textbf{Long Context} & \textbf{Method} & \multicolumn{1}{c}{\textbf{arXiv}} & \multicolumn{2}{c}{\textbf{Code}} & \multicolumn{1}{c}{\textbf{Code \&}} \\
      \textbf{Data} &  &  & Python & All & \textbf{arXiv}  \\
  
     \midrule
     \multirow{3}{*}{C} & \methodvariant{\bmtf{}} & \textbf{5.463} {\small $\pm$ .002} & \textbf{2.810} {\small $\pm$ .002} & \textbf{2.942} {\small $\pm$ .005} & \textbf{3.100} {\small $\pm$ .004} \\
     & \methodvariant{\contrieverms{}} & 5.477 {\small $\pm$ .005} & 2.824 $\pm$.001 & 2.957 {\small $\pm$ .006} & 3.115 {\small $\pm$ .006} \\
     & \methodvariant{\repo{}} & 5.474 {\small $\pm$ .007} & 2.827 {\small $\pm$ .006} & 2.958 {\small $\pm$ .009} & 3.115 {\small $\pm$ .009} \\
     & \textsc{EP} & 5.550 {\small $\pm$ .002} & 2.931 {\small $\pm$ .008} & 3.073 {\small $\pm$ .006} & 3.228 {\small $\pm$ .005} \\
      \bottomrule
    \end{tabular}
    \end{center}
    \caption{ Perplexity fine-tune on a $50/50$ data mixture of RedPajama and C code. We report the mean and {standard deviation}. Interestingly, training on the code data with \method{} improves general long-context performance on arXiv. 
    }\label{tab:cvariancecheck}

    \begin{center}
  \begin{tabular}{c c | c | c c c c | c}
  \toprule
  \textbf{Altered} & \textbf{Method} & \multicolumn{1}{c}{\textbf{arXiv}} & \multicolumn{4}{c}{\textbf{Code}} & \multicolumn{1}{c}{\textbf{Code \&}} \\
  \textbf{Data} &  &  & Haskell  & Python & CUDA & All & \textbf{arXiv}  \\
  \midrule
  \multirow{4}{*}{C\#}  
  & \methodvariant{\bmtf{}} & \textbf{4.86} \us{ .15} & \textbf{2.60} \us{ .19} & \textbf{2.66} \us{ .16} & \textbf{2.32} \us{ .19} & \textbf{2.75} \us{ .19} & \textbf{2.88} \us{ .19} \\
  & \methodvariant{\contrieverms{}} & 4.88 \us{ .13} & 2.62 \us{ .17} & 2.67 \us{ .15} & 2.34 \us{ .17} & 2.77 \us{ .17} & 2.90 \us{ .17} \\
  & \methodvariant{\repo{}} & 4.88 \us{ .13} & 2.62 \us{ .17} & 2.68 \us{ .14} & 2.35 \us{ .16} & 2.77 \us{ .17} & 2.90 \us{ .17} \\
   &   \textsc{EP} & 5.01 & 2.79 & 2.82 & 2.51 & 2.94 & 3.07 \\
  \bottomrule
  \end{tabular}
  \end{center}
  \caption{ Perplexity $_{(\text{imp over \textsc{EP}})}$ for training on a $50/50$ data mixture of RedPajama and C\# code with longer $64$K context.}\label{tab:csharpsek}

  \begin{center}
    \begin{tabular}{c |  c | c | c | c | c | c | c | c}
        \toprule
        \multicolumn{2}{c|}{Method} & \multicolumn{6}{c|}{\method{}} & \textsc{EP} \\
        \multicolumn{2}{c|}{Eval Data/Noise} & \textbf{0\%} & \textbf{10\%} & \textbf{25\%} & \textbf{50\%} & \textbf{75\%} & \textbf{90\%} & \textbf{-} \\
        \midrule
        \multicolumn{2}{c|}{\textbf{arXiv}} & \textbf{5.46} & 5.47 & 5.48 & 5.50 & 5.53 & 5.55 & 5.55 \\
        \multirow{2}{*}{\textbf{Code}} & Python & \textbf{2.81} & 2.82 & 2.83 & 2.86 & 2.89 & 2.92 & 2.93 \\
         & All & \textbf{2.94} & 2.95 & 2.97 & 3.01 & 3.04 & 3.06 & 3.07 \\
         \multicolumn{2}{c|}{\textbf{Code \& arXiv}} & \textbf{3.10} & 3.11 & 3.13 & 3.16 & 3.19 & 3.22 & 3.23 \\
        \bottomrule
    \end{tabular}
    \caption{We test the robustness of \method{} to noisy retriever. We achieve this by preparing data using \bmtf{} retriever that with probability $p$ returns a random document instead of the most related one. We note that \method{} is quite robust and only with $p=0.9$ approaches Example Packing.}
    \label{tab:data_noisy}
  \end{center}
  
  \end{table*}

  \begin{table}
    \centering
    \begin{tabular}{c c | c | c c  | c}
        \toprule
        \textbf{Altered} & \textbf{Method} & \multicolumn{1}{c}{\textbf{arXiv}} & \multicolumn{2}{c}{\textbf{Code}} & \multicolumn{1}{c}{\textbf{Code}} \\
        \textbf{Data} &  &  & Python  & All & \textbf{arXiv}  \\
        \midrule
        25\% &  \method{} & \textbf{5.43} &     \textbf{2.91} &        \textbf{3.08} &                  \textbf{3.22}  \\
        25\% &  \textsc{EP} &    5.49 &     3.00    &       3.19 &                  3.34 \\
        \midrule
        50\% &  \method{} & \textbf{5.46} & \textbf{2.81} & \textbf{2.94} & \textbf{3.10} \\
        50\% &  \textsc{EP} & 5.55 &     2.93 &        3.07 &                  3.23 \\
        \midrule
        75\% &  \method{} &  \textbf{5.61} &     \textbf{2.80}  &        \textbf{2.86} &                  \textbf{3.05} \\
        75\% &  \textsc{EP} & 5.73 &     2.94 &        3.04 &                 3.23 \\
        \bottomrule
        \end{tabular}
    \caption{We note that \method{} beats the \textsc{EP} perplexity when trained with various proportions of \methodvariant{\bmtf{}}/\textsc{EP} prepared C data (the remaining data is unaltered RedPajama).}
    \label{tab:data_proportions}
\end{table}

\subsection{Properties of \method{} Generated Data}\label{sec:splice_data_properties}
  \method{} conjecturally falls into the framework presented in \citep{data-distributional-properties}, which shows that the distributional properties of the training data affect the in-context capabilities of transformer models. In particular, it indicates the importance of  "burstiness", i.e., a flatter frequency distribution with a relatively higher mass on the rare, long-tail tokens appearing in a sequence. In Table \ref{tab:burstiness}, we show that \method{} increases the burstiness of the training data (measured in terms of Zipf's coefficient of token frequency) in comparison to the Example Packing.

  \begin{table}
    \
    \begin{center}
  \begin{tabular}{c | c | c }
  \toprule
  \textbf{Training Data} & \textbf{Method} & \textbf{Zipf's Coefficient} \\
  \midrule
  \multirow{2}{*}{{C}}  
  & \methodvariant{\bmtf{}} & ${1.512}_{(0.055)}$\\
  &     \textsc{EP} &${1.593}_{(0.025)}$\\
  
  \multirow{2}{*}{{StackEx}}   & \methodvariant{\bmtf{}} & ${1.643}_{( 0.026)}$\\
  &     \textsc{EP} &${1.664}_{(0.013)}$\\
  \bottomrule
  \end{tabular}
  \end{center}
  \caption{Zipf's coefficient of token frequency on \textsc{EP} and \method{} along with standard deviation. A lower Zipf's coefficient represents a more significant burstiness property.}
    \label{tab:burstiness}
  \end{table}

  \section{Related Work}\label{sec:related_work}
  There is an increasing number of works aiming to study the role of data in LLM training in detail. For instance,  \citep{levine2022icl} developed a theory and demonstrated empirically that incorporating non-adjacent but semantically related sentences in training samples leads to better sentence embeddings and improves open-domain question-answering performance. Another study by \citep{gu-etal-2023-pre} introduced a pretraining framework grounded on the idea that text documents often include intrinsic tasks. They showed that this approach substantially boosts in-context learning. Additionally, there is existing work on training long-context language models using repository-level code data, such as \citep{wu2022memorizing}. Work of \citep{data-distributional-properties} identifies the training data's distributional properties that affect transformer models' in-context capabilities. Similarly, \citep{han2023understanding} constructs small-scale data using an iterative gradient approach and shows that such data improve in-context performance.

  Our methodology diverges from these works in several key ways. First, while prior studies have focused on sentence-level \citep{levine2022icl} or paragraph-level \citep{gu-etal-2023-pre} granularity, we emphasize document-level context during training, specifically targeting long-context performance. We validate our approach in large-scale language modeling, using models such as OpenLLaMA $3$B, $7$B, and CodeLlama $13$B. Second, we construct a tree structure of related documents using BM25/Contriever-MSMARCO retrieval, which we then linearize to form long-context samples. This approach allows for greater control over the coherence of samples, compared to relying solely on natural data structures like repository-level code. While the gradient-based method in \citep{han2023understanding} shares similarities with our retrieval-based approach, our method scales to larger datasets and operates at a different granularity.

Concurrently with our research, \citep{shi2023incontext} introduced a method for preparing training data that shares similarities with \method{}, particularly in its default settings ($k=1$ with identity as \texttt{Order}). However, while their approach focuses on training models from scratch, our work demonstrates that long-context capabilities can be effectively achieved through short and cost-efficient fine-tuning.
In addition to this distinction, we employ significantly longer context lengths, extending above 64K tokens compared to the 8K tokens used in \citep{shi2023incontext}, which allows for more comprehensive context handling. Furthermore, we provide an in-depth analysis of our design choices, such as the advantages of data reordering (detailed in Appendix \apdx{sec:large_models_shuffling}{J}) and the impact of varying $k$ values (Appendix~\apdx{sec:appendix_hypparam_splice}{C.1}). These analyses underline the effectiveness and flexibility of our approach. Our findings are especially pertinent in the context of recent research on the "Physics of Language Models" \citep{allenzhu2024physicslanguagemodels31}, which discusses the limitations of fine-tuning. Despite these limitations, we show that \method{} offers substantial and quantifiable improvements even with relatively brief fine-tuning, providing a practical advantage in enhancing long-context capabilities.

  \section{Limitations and Future Work}\label{sec:limit_and_fut_work}

  We show that structuring the training data is a viable way of improving the model's long-context performance. The presented method, \method{}, can be viewed as a general framework for organizing the documents into training samples. This opens multiple further research avenues.

 \textbf{Retrieval Granularity} Another avenue for future work is to study the granularity of the pieces from which the training samples are constructed. In this work, we focus on the document-level granularity. However, it is possible to construct training samples from smaller pieces.

  \textbf{Other Data Sources} One of the approaches to training long-context language models is to use conversational data \citep{longchat2023}, which is complementary to our method. \method{} can utilize data that already exists in vast quantities and can be easily applied to different types of text (like code, Wikipedia articles, or StackExchange) to further increase the number of long-context samples. We leave researching how \method{} integrates with other methods for preparing the long-context data
  as future work.
  
  \textbf{Data Curation} Using highly correlated samples has the potential to result in training instability. However, we noted no performance degradation during our experiments. We leave the study of how \method{} integrates with different data types for the future. In particular, in our studies, the datasets used were reasonably deduplicated.

  \textbf{Neural Retriever} In our work, we have utilized Contriever \citep{izacard2022unsupervised} in a zero-shot setup, using the first $512$ tokens to generate the embedding. On the other hand, BM25 had access to all the document content. Further study is required to determine whether \method{} can additionally significantly benefit from properly tuned neural retrievers. In particular, in our case, Contriever tended to produce samples consisting of fewer repositories than \bmtf{}. We leave this for future work.

\section{Conclusions}
In this work, we present \method{}, a method of constructing training samples for long-context language models. It utilizes BM25/Contriever-MSMARCO to find relevant documents and feed them to the model in a structured manner. We show that \method{} improves performance on downstream tasks and the language modeling abilities of LLMs. We further show that \method{} can be used to improve long-context utilization of large-scale models using only short fine-tuning. We believe that our work indicates multiple interesting research directions for improving the performance of long-context language models with structured data.

\section*{Ethical Statement}
Our work develops a generic technique that allows for improving context utilization in language models via low-cost fine-tuning. However, we note that it does not create any new threats, but only exacerbates existing ones.
Therefore, we refer to the existing literature on the broader impact
of language models, such as \citep{borgeaud2022retro}.

\section*{Acknowledgments}
We are thankful for the TPU Research Cloud program, which was instrumental to our
research by providing significant computational resources. Parts of the project were realized using
the resources of IDEAS NCBR.

\small{\bibliography{aaai25}}

\newpage

\appendix
\onecolumn
\twocolumn

\section{Architecture}\label{sec:appendix_architecture}
The architecture of our models is based on LLaMA \citep{touvron2023llama}, and the architectural details can be found in Table \ref{tab:arch_details}. Briefly speaking, our architecture is similar to the one introduced in \citep{vaswani2017attention} with a few standard changes. First, we use only the decoder without the encoder part. Secondly, we perform RMSNorm before the input of both the attention and feed-forward modules. Thirdly, we use the LLaMA FeedForward module. Additionally, we use Rotary Position Embedding \citep{su2021roformer}. 
For context extension, we use Focused Transformer (FoT)\citep{tworkowski2023focused}, CodeLlama \citep{roziere2023code} (CL) and YaRN \citep{peng2023yarn}.
Table \ref{tab:train_details} presents the details about both standard and long-context pretraining/fine-tuning. We use AdamW as an optimizer, with $\beta_1=0.9$ and $\beta_2=0.95$.

\begin{table*}

    \begin{center}
    \begin{tabular}{l | c c c c}
        \toprule
        \textbf{Parameter/Model Size} & \textbf{270M} & \textbf{3B} & \textbf{7B} & \textbf{13B} \\
        \midrule
        Vocab Size & 32000 & 32000 & 32000 & 32016 \\
        Embedding Size & 1024 & 3200 & 4096 & 5120 \\
        Num Attention Layers & 12 & 26 & 32 & 40 \\
        Num Attention Heads & 8 & 32 & 32 & 40\\
        Head Size & 128 & 100 & 128 & 128 \\
        MLP Hidden Size & 4096 & 8640 & 11008 & 13824 \\
        FoT Context Extension Layers & [6, 9] & [6, 12, 18] & [8, 16, 24] & [12, 24, 36]  \\
        \hline
    \end{tabular}
    
    \end{center}
    \caption{\small Architecture details. Focused Transformer context extension is applied in fine-tuning for $32$K context and evaluation. }
    \label{tab:arch_details}
\end{table*}

Each tuning experiment of 270M model was done using either a TPUv3-8 or TPUv4-8 machine and took around $10$ hours. 
Tuning a $3$B parameter model for $5.4$B tokens with FoT took $40$ hours on TPUv3-128.
The $7$B and $13$B models were tuned on TPUv3-128.

\begin{table*}

    \begin{center}
    \begin{tabular}{l l | c c c c}
        \toprule
        \textbf{Stage} & \textbf{Parameter/Model Size} & \textbf{270M} & \textbf{3B} & \textbf{7B} & \textbf{13B} \\
        \midrule
        \multirow{2}{*}{Pretraining} & Context & 2K & 2K & 2K & 16K \\
         & Tokens & 6.3B & 1T & 1T & 2.5T \\
         \midrule
        \multirow{6}{*}{\begin{tabular}{@{}c@{}}\small Long \\ Context \\ Tuning \end{tabular} } & Context & 32K\textsubscript{\tiny FoT}, 16K\textsubscript{\tiny no-FoT} & 32K & 32K & 32K \\
         & Batch Size & 128\textsubscript{\tiny FoT}, 16\textsubscript{\tiny no-FoT} & 128 & 128\textsubscript{\tiny FoT}, 32\textsubscript{\tiny no-FoT} & 128 \\
         & Start Learning Rate & 5e-5 & 1.5e-5 & 1.5e-5 & 1.5e-5  \\
         & End Learning Rate & 5e-6 & 1.5e-6 & 1.5e-6 & 1.5e-6 \\
         & Warmup Steps & 250 & 1000 & 1000\textsubscript{\tiny FoT}, 200\textsubscript{\tiny no-FoT} & 1000  \\
         & Tokens & 1B & {5.4B} & 2B & 2B  \\
        \hline
    \end{tabular}
    
    \end{center}
    \caption{\small Training details. We pretrain a custom $270$M parameter model and take a pretrained $3$B/$7$B/ parameter OpenLLaMAv2 model \citep{geng2023easylm} and $13$B CodeLlama \citep{roziere2023code} model. Subscript denotes that parameter was specific for a context extension method with FoT referring to \citep{tworkowski2023focused} and no-FoT to other methods (Naive, YaRN \citep{peng2023yarn}, CodeLlama \citep{roziere2023code}).}
    \label{tab:train_details}
\end{table*}



\section{Additional Results for $270$M Models}

\subsection{Short Context Evaluation}\label{sec:short_context_270}
In Table \ref{tab:medium_short_ctx}, we asses the short context performance of $270$M models from Section \ref{sec:exps_medium_scale} and compare them against their starting checkpoint.
\begin{table*}
  \centering
  
    \begin{tabular}{ll|c|cccc|c}
    \toprule
    \multirow{2}{*}{Tune Dataset} & \textbf{Method} & \multicolumn{1}{c}{\textbf{arXiv}} &  \multicolumn{4}{c}{\textbf{Code}} & \multicolumn{1}{c}{\textbf{Code \&}} \\
    &  &  & Haskell & Python & CUDA & All & \textbf{arXiv} \\
    \midrule
      & \textsc{ST CHKPT} & 10.80 & 13.40 & 6.00 & 5.00 & 7.17 & 7.40 \\
    \midrule
    \multirow{3}{*}{Wikipedia}& \methodvariant{\bmtf{}} & 10.64 & 13.28 & 6.00 & 5.00 & 7.16 & 7.38 \\
     & \methodvariant{\contrieverms{}} & 10.60 & 13.24 & 5.99 & 4.99 & 7.14 & 7.36 \\
     & \textsc{EP} & 10.59 & 13.20 & 5.99 & 5.00 & 7.12 & 7.34 \\
    \midrule
    \multirow{3}{*}{Stackexchange} & \methodvariant{\bmtf{}} & 9.28 & 12.94 & 5.91 & 4.89 & 7.05 & 7.19 \\
     & \methodvariant{\contrieverms{}} & 9.26 & 12.88 & 5.89 & 4.88 & 7.04 & 7.18 \\
     & \textsc{EP} & 9.24 & 13.00 & 5.91 & 4.90 & 7.07 & 7.21 \\
     \midrule
     \multirow{4}{*}{C}     & \methodvariant{\bmtf{}} & 10.76 & 12.10 & 5.55 & 3.78 & 6.24 & 6.52 \\
         & \methodvariant{\contrieverms{}} & 10.76 & 12.06 & 5.53 & 3.76 & 6.21 & 6.50 \\
         & \textsc{EP} & 10.73 & 11.96 & 5.49 & 3.68 & 6.16 & 6.44 \\
         & \methodvariant{\repo{}} & 10.78 & 12.10 & 5.56 & 3.79 & 6.24 & 6.52 \\
         \midrule
    \multirow{4}{*}{C\#}   & \methodvariant{\bmtf{}} & 10.86 & 12.74 & 5.73 & 4.67 & 6.72 & 6.98 \\
       & \methodvariant{\contrieverms{}} & 10.89 & 12.72 & 5.72 & 4.65 & 6.70 & 6.96 \\
       & \textsc{EP} & 10.83 & 12.73 & 5.70 & 4.65 & 6.69 & 6.95 \\
       & \methodvariant{\repo{}} & 10.91 & 12.77 & 5.74 & 4.68 & 6.72 & 6.98 \\
       \midrule
    \multirow{4}{*}{Python}  & \methodvariant{\bmtf{}} & 10.75 & 12.33 & 4.12 & 4.44 & 6.38 & 6.65 \\
     & \methodvariant{\contrieverms{}} & 10.76 & 12.28 & 4.09 & 4.43 & 6.36 & 6.64 \\
     & \textsc{EP} & 10.72 & 12.19 & 3.98 & 4.40 & 6.33 & 6.61 \\
     & \methodvariant{\repo{}} & 10.78 & 12.33 & 4.12 & 4.45 & 6.38 & 6.66 \\
    \bottomrule
    \end{tabular}%
    \caption{$2$K context perplexity evaluation of models from Section \ref{sec:exps_medium_scale}}\label{tab:medium_short_ctx}%
\end{table*}%

\subsection{Different Context Extension Methods }\label{sec:med_diff_ctx}
We test \method{} with different context extension methods. In Table \ref{tab:show_on_more_ctx_ext_methods}, we show that \method{} brings improvements also when context is extended in all layers using the naive approach, the CodeLlama \citep{roziere2023code}, and \citep{peng2023yarn} RoPe \citep{su2021roformer} adjustment method.

\begin{table*}[h]

 \begin{center}
 \vspace{-0.15cm}
 \begin{tabular}{c  c | c | c c c c | c}
   \toprule
   \textbf{RoPe}  & \textbf{Method} & \multicolumn{1}{c}{\textbf{arXiv}} & \multicolumn{4}{c}{\textbf{Code}} & \multicolumn{1}{c}{\textbf{Code \&}} \\
   \textbf{scale}  &  &  & Haskell  & Python & CUDA & All & \textbf{arXiv}  \\

   \midrule
   \multirow{3}{*}{{Naive}}
 
   & \methodvariant{\bmtf{}} & \textbf{6.25} \us{ .08} & \textbf{4.84} \us{ .19} & \textbf{3.55} \us{ .11} & \textbf{2.84} \us{ .12} & \textbf{3.72} \us{ .13} & \textbf{3.88} \us{ .12} \\
   &  \methodvariant{\repo{}} & \textbf{6.25} \us{ .08} & 4.87 \us{ .16} & 3.56 \us{ .10} & 2.85 \us{ .11} & 3.74 \us{ .11} & 3.89 \us{ .11} \\
   &  \textsc{EP} & 6.33 & 5.03 & 3.66 & 2.96 & 3.85 & 4.00 \\
   \midrule
   \multirow{3}{*}{{CL}}  
   & \methodvariant{\bmtf{}} & \textbf{5.74} \us{ .02} & \textbf{4.28} \us{ .09} & \textbf{3.22} \us{ .05} & \textbf{2.53} \us{ .05} & \textbf{3.34} \us{ .06} & \textbf{3.49} \us{ .06} \\
   &  \methodvariant{\repo{}} & \textbf{5.74} \us{ .02} & \textbf{4.28} \us{ .09} & \textbf{3.22} \us{ .05} & 2.54 \us{ .04} & 3.35 \us{ .05} & 3.50 \us{ .05} \\
   &  \textsc{EP} & 5.76 & 4.37 & 3.27 & 2.58 & 3.40 & 3.55 \\
   \midrule
   \multirow{3}{*}{{YaRN}} 
   & \methodvariant{\bmtf{}} & \textbf{5.77} \us{ .02} & \textbf{4.32} \us{ .10} & \textbf{3.24} \us{ .05} & \textbf{2.55} \us{ .06} & \textbf{3.37} \us{ .07} & \textbf{3.52} \us{ .06} \\
&  \methodvariant{\repo{}} & \textbf{5.77} \us{ .02} & \textbf{4.32} \us{ .10} & \textbf{3.24} \us{ .05} & 2.56 \us{ .05} & 3.38 \us{ .06} & 3.53 \us{ .05} \\
   &  \textsc{EP} & 5.79 & 4.42 & 3.29 & 2.61 & 3.44 & 3.58 \\
   \bottomrule
 \end{tabular}
 \end{center}
 \caption{\small Perplexity $_{(\text{imporovment over \textsc{EP}})}$ for training on a $50/50$ data mixture of RedPajama, and C\#.
  We check that  \method{} brings improvements when fine-tuning for the longer context ($16$K)  using the method of CodeLlama \citep{roziere2023code}, YaRN \citep{peng2023yarn}, or left without changes (Naive), as opposed to FoT used in the other experiments. For details see Table \ref{tab:bigresultsclyarn}.} \label{tab:show_on_more_ctx_ext_methods}
\end{table*}

\subsection{Detailed Results}\label{sec:detailed_results}
In this section, we extend results presented in Section \ref{sec:exps_medium_scale}. Details about the number of tokens used for evaluation can be found in Table \ref{tab:eval_tokens}.

Tables \ref{tab:bigresultsccspy} and \ref{tab:big_cvar}  show the results of training $270$M parameter model for $32$K context on a 50/50 mixture of RedPajama data (organized in a standard way) and code data organized using a specified method.
Table \ref{tab:bigresultsccspy} contains detailed results from training on C\# and Python. Table \ref{tab:big_cvar} contains results on C averaged across three different subsets of C (for details about the construction of those subsets, see Appendix \ref{sec:apendix_data_prep}). Both tables show that \method{} outperforms the \textsc{EP} by a significant margin.

The main advantage of \methodvariant{\bmtf{}}/\methodvariant{\contrieverms{}{}} over the \methodvariant{\repo{}} approach is that it can also be used for non-structured data. Table \ref{tab:bigresultsstack} shows the detailed results of applying the \method{} on non-code data. Note that training on non-code data allows us to improve the model perplexity on the arXiv dataset in comparison to the model trained on code.

Tables \ref{tab:bigresultsclyarn}, \ref{tab:bigresultsclnaive} consider models trained with YaRN \citep{peng2023yarn}, CodeLlama \citep{roziere2023code} and Naive (no adjustment to RoPE) context extension methods.
Table \ref{tab:bigresultsrandbasreposplice} shows that a simple artificial extension of example length via random concatenation of documents within domain (C source files) does not help.

\begin{table*}

  \begin{center}
\begin{tabular}{lcccc}
\toprule

\multicolumn{5}{c}{Altered Train Data: C}  \\
 Eval/Method & 
\methodvariant{\bmtf{}} & 
\methodvariant{\contrieverms{}} & 
\textsc{EP} & 
\methodvariant{\repo{}} \\
\midrule
ArXiv & \textbf{5.463} $\pm$ 0.002 & 5.477 $\pm$ 0.005 & 5.550 $\pm$ 0.002 & 5.474 $\pm$ 0.007 \\
C & \textbf{2.126} $\pm$ 0.020 & 2.134 $\pm$ 0.020 & 2.173 $\pm$ 0.023 & 2.135 $\pm$ 0.020 \\
C++ & \textbf{2.396} $\pm$ 0.004 & 2.403 $\pm$ 0.002 & 2.467 $\pm$ 0.001 & 2.403 $\pm$ 0.006 \\
CUDA & \textbf{2.219} $\pm$ 0.002 & 2.235 $\pm$ 0.005 & 2.330 $\pm$ 0.009 & 2.239 $\pm$ 0.004 \\
C\# & \textbf{1.939} $\pm$ 0.000 & 1.948 $\pm$ 0.001 & 2.016 $\pm$ 0.004 & 1.949 $\pm$ 0.002 \\
Common Lisp & \textbf{2.994} $\pm$ 0.072 & 3.042 $\pm$ 0.091 & 3.195 $\pm$ 0.078 & 3.043 $\pm$ 0.102 \\
Dart & \textbf{2.141} $\pm$ 0.002 & 2.155 $\pm$ 0.004 & 2.268 $\pm$ 0.009 & 2.156 $\pm$ 0.002 \\
Emacs Lisp & 7.857 $\pm$ 0.017 & 7.851 $\pm$ 0.020 & 8.098 $\pm$ 0.027 & \textbf{7.840} $\pm$ 0.019 \\
Erlang & \textbf{2.665} $\pm$ 0.002 & 2.680 $\pm$ 0.003 & 2.769 $\pm$ 0.007 & 2.676 $\pm$ 0.003 \\
Fortran & \textbf{3.306} $\pm$ 0.010 & 3.335 $\pm$ 0.010 & 3.554 $\pm$ 0.010 & 3.333 $\pm$ 0.012 \\
Go & \textbf{1.910} $\pm$ 0.001 & 1.924 $\pm$ 0.007 & 1.999 $\pm$ 0.002 & 1.924 $\pm$ 0.006 \\
Groovy & \textbf{3.009} $\pm$ 0.001 & 3.026 $\pm$ 0.006 & 3.147 $\pm$ 0.013 & 3.025 $\pm$ 0.007 \\
Haskell & \textbf{3.198} $\pm$ 0.001 & 3.221 $\pm$ 0.001 & 3.371 $\pm$ 0.008 & 3.220 $\pm$ 0.008 \\
Java & \textbf{2.075} $\pm$ 0.002 & 2.086 $\pm$ 0.001 & 2.161 $\pm$ 0.006 & 2.085 $\pm$ 0.005 \\
Pascal & \textbf{3.492} $\pm$ 0.026 & 3.496 $\pm$ 0.019 & 3.622 $\pm$ 0.021 & 3.513 $\pm$ 0.014 \\
Python & \textbf{2.810} $\pm$ 0.002 & 2.824 $\pm$ 0.001 & 2.931 $\pm$ 0.008 & 2.827 $\pm$ 0.006 \\
\midrule
Mean & \textbf{3.100} $\pm$ 0.004 & 3.115 $\pm$ 0.006 & 3.228 $\pm$ 0.005 &  3.115 $\pm$ 0.009 \\
\bottomrule
\end{tabular}

\end{center}
\caption{\small To check the statistical significance of our results, we prepare three subsets of C (details in Appendix \ref{sec:apendix_data_prep}) and train the models on a 50/50 mixture of RedPajama data (organized in the standard way) and C data organized using one of the methods. Note that the standard deviation is much lower than the perplexity improvements from using \method{}.}
\label{tab:big_cvar}
\end{table*}

\begin{table*}
  \begin{center}

\begin{tabular}{lcccc}
\toprule
\multicolumn{5}{c}{Altered Train Data: C\#}  \\
 Eval/Method & 
\methodvariant{\bmtf{}} & 
\methodvariant{\contrieverms{}} & 
\textsc{EP} & 
\methodvariant{\repo{}} \\
\midrule
ArXiv & \textbf{5.52} & 5.53 & 5.65 & 5.53 \\
C & \textbf{2.39} & 2.40 & 2.50 & 2.40 \\
C++ & \textbf{2.60} & 2.61 & 2.74 & 2.62 \\
CUDA & \textbf{2.46} & 2.48 & 2.65 & 2.49 \\
C\# & \textbf{1.82} & \textbf{1.82} & 1.90 & \textbf{1.82} \\
Common Lisp & \textbf{3.41} & 3.44 & 3.72 & 3.42 \\
Dart & \textbf{2.14} & 2.16 & 2.31 & 2.16 \\
Emacs Lisp & \textbf{8.41} & 8.46 & 8.85 & \textbf{8.41} \\
Erlang & \textbf{2.80} & \textbf{2.80} & 2.95 & 2.81 \\
Fortran & \textbf{3.68} & 3.71 & 4.05 & 3.72 \\
Go & \textbf{1.94} & 1.95 & 2.06 & 1.96 \\
Groovy & \textbf{3.01} & 3.03 & 3.25 & 3.04 \\
Haskell & \textbf{3.33} & 3.35 & 3.58 & 3.35 \\
Java & \textbf{2.09} & 2.10 & 2.22 & 2.10 \\
Pascal & 3.61 & 3.62 & 3.80 & \textbf{3.60} \\
Python & \textbf{2.90} & 2.91 & 3.07 & 2.91 \\
\midrule
Mean & \textbf{3.26} & 3.27 & 3.46 & 3.27 \\
\bottomrule
\end{tabular}

\begin{tabular}{lcccc}
\toprule

\multicolumn{5}{c}{Altered Train Data: Python}  \\
 Eval/Method & 
\methodvariant{\bmtf{}} & 
\methodvariant{\contrieverms{}} & 
\textsc{EP} & 
\methodvariant{\repo{}} \\
\midrule
ArXiv & \textbf{5.47} & 5.49 & 5.57 & 5.48 \\
C & \textbf{2.38} & 2.39 & 2.45 & 2.39 \\
C++ & \textbf{2.61} & 2.63 & 2.71 & 2.63 \\
CUDA & \textbf{2.41} & 2.43 & 2.56 & 2.44 \\
C\# & \textbf{1.98} & 1.99 & 2.06 & 2.00 \\
Common Lisp & \textbf{3.23} & 3.28 & 3.47 & 3.27 \\
Dart & \textbf{2.15} & 2.17 & 2.27 & 2.17 \\
Emacs Lisp & 7.94 & 7.98 & 8.28 & \textbf{7.93} \\
Erlang & \textbf{2.70} & 2.71 & 2.82 & 2.71 \\
Fortran & \textbf{3.42} & 3.46 & 3.72 & 3.46 \\
Go & \textbf{1.95} & 1.96 & 2.03 & 1.96 \\
Groovy & \textbf{2.97} & 2.99 & 3.13 & 2.99 \\
Haskell & \textbf{3.25} & 3.28 & 3.46 & 3.27 \\
Java & \textbf{2.12} & \textbf{2.12} & 2.20 & 2.13 \\
Pascal & \textbf{3.57} & 3.58 & 3.73 & 3.58 \\
Python & \textbf{2.53} & \textbf{2.53} & 2.62 & 2.54 \\
\midrule
Mean & \textbf{3.17} & 3.19 & 3.32 & 3.18 \\
\bottomrule
\end{tabular}


\end{center}
\caption{\small Perplexity results comparing different ways of organizing the same data.
  All runs started from the same $270M$ model with $2048$ context and were trained for $32$K context on a 50/50 mixture of RedPajama 
  (organized in a standard way) and code organized in the mentioned ways. For details about training, please refer to Section \ref{sec:exps_medium_scale}.}
  \label{tab:bigresultsccspy}
\end{table*}

\begin{table}

  \begin{center}

\begin{tabular}{lccc}
\toprule
\multicolumn{4}{c}{Altered Train Data: StackExchange}  \\
Eval/Method & 
\methodvariant{\bmtf{}} & 
\methodvariant{\contrieverms{}} & 
\textsc{EP} \\
\midrule
ArXiv & \textbf{5.07} & 5.09 & 5.14 \\
C & \textbf{2.68} & 2.69 & 2.70 \\
C++ & \textbf{3.02} & 3.04 & 3.06 \\
CUDA & \textbf{2.89} & 2.93 & 2.94 \\
C\# & \textbf{2.27} & 2.28 & 2.29 \\
Common Lisp & \textbf{4.02} & 4.06 & 4.08 \\
Dart & \textbf{2.58} & 2.60 & 2.61 \\
Emacs Lisp & \textbf{9.55} & 9.67 & 9.69 \\
Erlang & \textbf{3.13} & 3.16 & 3.18 \\
Fortran & \textbf{4.28} & 4.34 & 4.38 \\
Go & \textbf{2.24} & 2.25 & 2.27 \\
Groovy & \textbf{3.62} & 3.66 & 3.68 \\
Haskell & \textbf{3.88} & 3.91 & 3.94 \\
Java & \textbf{2.43} & 2.45 & 2.45 \\
Pascal & \textbf{4.08} & 4.11 & 4.14 \\
Python & \textbf{3.32} & 3.35 & 3.36 \\
\midrule
Mean & \textbf{3.69} & 3.73 & 3.74 \\
\bottomrule
\end{tabular}

\begin{tabular}{lccc}
\toprule
\multicolumn{4}{c}{Altered Train Data: Wikipedia}  \\
 Eval/Method & 
\methodvariant{\bmtf{}} & 
\methodvariant{\contrieverms{}} & 
\textsc{EP} \\
\midrule
ArXiv & \textbf{5.64} & 5.65 & 5.73 \\
C & \textbf{2.65} & 2.67 & 2.71 \\
C++ & \textbf{2.98} & 3.01 & 3.07 \\
CUDA & \textbf{2.87} & 2.92 & 3.00 \\
C\# & \textbf{2.22} & 2.24 & 2.29 \\
Common Lisp & \textbf{3.87} & 3.96 & 4.08 \\
Dart & \textbf{2.51} & 2.55 & 2.61 \\
Emacs Lisp & \textbf{9.38} & 9.45 & 9.63 \\
Erlang & \textbf{3.13} & 3.16 & 3.23 \\
Fortran & \textbf{4.23} & 4.32 & 4.49 \\
Go & \textbf{2.18} & 2.21 & 2.26 \\
Groovy & \textbf{3.49} & 3.55 & 3.67 \\
Haskell & \textbf{3.82} & 3.87 & 3.97 \\
Java & \textbf{2.39} & 2.41 & 2.46 \\
Pascal & 4.32 & \textbf{4.23} & 4.40 \\
Python & \textbf{3.26} & 3.30 & 3.37 \\
\midrule
Mean & \textbf{3.68} & 3.72 & 3.81 \\
\bottomrule
\end{tabular}

\end{center}
\caption{\small Perplexity results comparing different ways of organizing the same data.
All runs started from the same $270M$ model with $2048$ context and were trained for $32$K context on a 50/50 mixture of RedPajama 
(organized in a standard way) and other data organized using one of the methods. For details about training, please refer to Section \ref{sec:exps_medium_scale}.
Note that the model trained with \method{} on StackExchange outperforms the one trained on code on arXiv evaluation, showing the benefits of \method{}'s applicability to non-code data.}
\label{tab:bigresultsstack}
\end{table}

\begin{table}

  \begin{center}
\begin{tabular}{lccc}
\toprule
 \multicolumn{4}{c}{Altered Train Data: C\#}  \\
\multicolumn{4}{c}{Context $16$K: CodeLlama}\\
Eval/Method & 
\methodvariant{\bmtf{}} & 
\textsc{EP} & 
\methodvariant{\repo{}} \\
\midrule
ArXiv & \textbf{5.74} & 5.76 & \textbf{5.74} \\
C & \textbf{2.66} & 2.70 & \textbf{2.66} \\
C++ & \textbf{2.79} & 2.83 & \textbf{2.79} \\
CUDA & \textbf{2.53} & 2.58 & 2.54 \\
C\# & \textbf{1.91} & 1.93 & \textbf{1.91} \\
Common Lisp & \textbf{3.78} & 3.85 & 3.79 \\
Dart & \textbf{2.28} & 2.33 & \textbf{2.28} \\
Emacs Lisp & \textbf{8.29} & 8.41 & 8.30 \\
Erlang & \textbf{3.57} & 3.64 & 3.58 \\
Fortran & \textbf{3.93} & 4.01 & 3.95 \\
Go & \textbf{1.99} & 2.03 & 2.00 \\
Groovy & \textbf{2.95} & 3.01 & 2.96 \\
Haskell & \textbf{4.28} & 4.37 & \textbf{4.28} \\
Java & \textbf{2.31} & 2.35 & \textbf{2.31} \\
Pascal & \textbf{3.67} & 3.72 & \textbf{3.67} \\
Python & \textbf{3.22} & 3.27 & \textbf{3.22} \\
\midrule
Mean & \textbf{3.49} & 3.55 & 3.50 \\
\bottomrule
\end{tabular}

\begin{tabular}{lccc}
\toprule
\multicolumn{4}{c}{Context $16$K: YaRN}\\

Eval/Method & 
\methodvariant{\bmtf{}} &
\textsc{EP} & 
\methodvariant{\repo{}} \\
\midrule
ArXiv & \textbf{5.77} & 5.79 & \textbf{5.77} \\
C & \textbf{2.68} & 2.72 & \textbf{2.68} \\
C++ & \textbf{2.81} & 2.85 & \textbf{2.81} \\
CUDA & \textbf{2.55} & 2.61 & 2.56 \\
C\# & \textbf{1.92} & 1.94 & \textbf{1.92} \\
Common Lisp & \textbf{3.83} & 3.92 & 3.84 \\
Dart & \textbf{2.30} & 2.36 & \textbf{2.30} \\
Emacs Lisp & \textbf{8.37} & 8.52 & 8.38 \\
Erlang & \textbf{3.60} & 3.67 & 3.61 \\
Fortran & \textbf{3.97} & 4.06 & 3.99 \\
Go & \textbf{2.00} & 2.04 & 2.01 \\
Groovy & \textbf{2.98} & 3.03 & \textbf{2.98} \\
Haskell & \textbf{4.32} & 4.42 & \textbf{4.32} \\
Java & \textbf{2.33} & 2.37 & \textbf{2.33} \\
Pascal & \textbf{3.69} & 3.75 & \textbf{3.69} \\
Python & \textbf{3.24} & 3.29 & \textbf{3.24} \\
\midrule
Mean & \textbf{3.52} & 3.58 & 3.53 \\
\bottomrule
\end{tabular}

\end{center}
\caption{\small Perplexity results comparing different ways of organizing the same data for non-FoT models.
All runs started from the same $270M$ model with $2048$ context and were trained for $16$K context on a 50/50 mixture of RedPajama 
(organized in a standard way) and C\# code is organized in one of three ways.}
\label{tab:bigresultsclyarn}
\end{table}

\begin{table}

\begin{center}
\begin{tabular}{lccc}
\toprule
\multicolumn{4}{c}{Altered Train Data: C\#}  \\
\multicolumn{4}{c}{Context $16$K: Naive}\\
 Eval/Method & \methodvariant{\bmtf{}} &
\textsc{EP} & 
\methodvariant{\repo{}} \\
\midrule
ArXiv & \textbf{6.25} & 6.33 & \textbf{6.25} \\
C & \textbf{2.88} & 2.96 & 2.89 \\
C++ & \textbf{3.04} & 3.13 & 3.05 \\
CUDA & \textbf{2.84} & 2.96 & 2.85 \\
C\# & \textbf{2.04} & 2.08 & \textbf{2.04} \\
Common Lisp & 4.40 & 4.56 & \textbf{4.39} \\
Dart & \textbf{2.50} & 2.60 & 2.51 \\
Emacs Lisp & \textbf{9.25} & 9.46 & \textbf{9.25} \\
Erlang & \textbf{3.98} & 4.10 & 4.00 \\
Fortran & \textbf{4.56} & 4.79 & 4.59 \\
Go & \textbf{2.14} & 2.21 & 2.16 \\
Groovy & \textbf{3.27} & 3.39 & 3.28 \\
Haskell & \textbf{4.84} & 5.03 & 4.87 \\
Java & \textbf{2.52} & 2.60 & 2.53 \\
Pascal & \textbf{4.05} & 4.20 & 4.10 \\
Python & \textbf{3.55} & 3.66 & 3.56 \\
\midrule
Mean & \textbf{3.88} & 4.00 & 3.89 \\
\bottomrule
\end{tabular}
\caption{\small Perplexity results comparing different ways of organizing the same data for non-FoT models.
  All runs started from the same $270M$ model with $2048$ context and were trained for $16$K context on a 50/50 mixture of RedPajama with a Naive context extension method.}\label{tab:bigresultsclnaive}

  \begin{tabular}{llll}
\toprule
 & 
\methodvariant{\bmtf{}} & 
\methodvariant{\contrieverms{}} & 
\textsc{Baseline} \\
Eval/Method &  &  &  \\
\midrule
ArXiv & \textbf{4.55} & 4.56 & 4.71 \\
C & \textbf{2.02} & 2.03 & 2.12 \\
C++ & \textbf{1.99} & \textbf{1.99} & 2.13 \\
CUDA & \textbf{1.97} & 1.98 & 2.14 \\
C\# & \textbf{1.58} & \textbf{1.58} & 1.66 \\
Common Lisp & \textbf{2.83} & 2.84 & 3.08 \\
Dart & \textbf{2.13} & 2.14 & 2.28 \\
Emacs Lisp & \textbf{5.90} & 5.92 & 6.26 \\
Erlang & \textbf{1.73} & 1.74 & 1.83 \\
Fortran & \textbf{2.75} & 2.76 & 3.06 \\
Go & \textbf{1.76} & \textbf{1.76} & 1.87 \\
Groovy & \textbf{2.32} & \textbf{2.32} & 2.41 \\
Haskell & \textbf{2.13} & 2.14 & 2.28 \\
Java & \textbf{1.68} & 1.69 & 1.81 \\
Pascal & 3.82 & \textbf{3.81} & 4.05 \\
Python & \textbf{2.49} & 2.51 & 2.63 \\
\midrule
Mean & \textbf{2.60} & 2.61 & 2.77 \\
\bottomrule
\end{tabular}
\caption{\small Perplexity for training on a $50/50$ data mixture of RedPajama and C\# code with longer $131$K context. We note that \method{} still brings significant benefits here, in particular when compared with $32$K and $64$K setups.}\label{tab:csharpsek128}
\end{center}

\end{table}

\begin{table}
  
  \begin{center}
\begin{tabular}{llll}
\toprule
 & 
\methodvariant{\bmtf{}} & 
\methodvariant{\contrieverms{}} & 
\textsc{Baseline} \\
Eval/Method &  &  &  \\
\midrule
ArXiv & \textbf{4.36} & \textbf{4.36} & 4.51 \\
Python & \textbf{2.73} & 2.74 & 2.85 \\
\midrule
Mean & \textbf{3.54} & 3.55 & 3.68 \\
\bottomrule
\end{tabular}
\end{center}
\caption{\small Perplexity for training on a $50/50$ data mixture of RedPajama and C\# code with longer $131$K context and evaluating on $160$K context. We note that FoT uses positional encodings that allow for such extrapolation.}\label{tab:csharpsek128160}
\end{table}

\begin{table}

  \begin{center}
\begin{tabular}{lccc}
\toprule
 \multicolumn{4}{c}{Altered Train Data: C }  \\
Eval/Method & 
\methodvariant{\bmtf{}} & 
\textsc{EP} &
\textsc{WithinDomEP} \\

\midrule
ArXiv & \textbf{5.46} & 5.55 & 5.55 \\
C & \textbf{2.13} & 2.17 & 2.18 \\
C++ & \textbf{2.40} & 2.47 & 2.47 \\
CUDA & \textbf{2.22} & 2.33 & 2.33 \\
C\# & \textbf{1.94} & 2.02 & 2.02 \\
Common Lisp & \textbf{2.99} & 3.20 & 3.18 \\
Dart & \textbf{2.14} & 2.27 & 2.27 \\
Emacs Lisp & \textbf{7.86} & 8.10 & 8.09 \\
Erlang & \textbf{2.67} & 2.77 & 2.77 \\
Fortran & \textbf{3.31} & 3.55 & 3.56 \\
Go & \textbf{1.91} & 2.00 & 2.00 \\
Groovy & \textbf{3.01} & 3.15 & 3.15 \\
Haskell & \textbf{3.20} & 3.37 & 3.37 \\
Java & \textbf{2.07} & 2.16 & 2.16 \\
Pascal & \textbf{3.49} & 3.62 & 3.64 \\
Python & \textbf{2.81} & 2.93 & 2.93 \\
\midrule
Mean & \textbf{3.10} & 3.23 & 3.23 \\
\bottomrule
\end{tabular}

\end{center}

\caption{\small Perplexity results comparing different ways of organizing the same data.
All runs started from the same $270M$ model with $2048$ context and were trained for $32$K context on a 50/50 mixture of RedPajama 
(organized in a standard way) and C code is organized in one of three ways.}
\label{tab:bigresultsrandbasreposplice}
\end{table}

\newpage

\onecolumn

\begin{table}
    \begin{center}

\begin{tabular}{lllll}
\toprule
 & 
\methodvariant{\bmtf{}} & 
\methodvariant{\contrieverms{}} & 
\textsc{Baseline} & 
\textsc{Repo} \\
Eval/Method &  &  &  &  \\
\midrule
ArXiv & \textbf{4.86} & 4.88 & 5.01 & 4.88 \\
C & \textbf{2.17} & 2.19 & 2.28 & 2.19 \\
C++ & \textbf{2.29} & 2.30 & 2.43 & 2.30 \\
CUDA & \textbf{2.32} & 2.34 & 2.51 & 2.35 \\
C\# & \textbf{1.72} & 1.73 & 1.81 & 1.73 \\
Common Lisp & \textbf{2.90} & 2.95 & 3.15 & 2.96 \\
Dart & \textbf{2.07} & 2.09 & 2.22 & 2.09 \\
Emacs Lisp & \textbf{7.40} & 7.43 & 7.84 & 7.42 \\
Erlang & \textbf{2.10} & 2.11 & 2.23 & 2.11 \\
Fortran & \textbf{3.15} & 3.19 & 3.50 & 3.20 \\
Go & \textbf{1.82} & 1.83 & 1.92 & 1.83 \\
Groovy & \textbf{2.54} & 2.59 & 2.76 & 2.58 \\
Haskell & \textbf{2.60} & 2.62 & 2.79 & 2.62 \\
Java & \textbf{1.83} & 1.85 & 1.96 & 1.84 \\
Pascal & \textbf{3.62} & 3.64 & 3.82 & \textbf{3.62} \\
Python & \textbf{2.66} & 2.67 & 2.82 & 2.68 \\
\midrule
Mean & \textbf{2.88} & 2.90 & 3.07 & 2.90 \\
\bottomrule
\end{tabular}
\caption{\small Perplexity $_{(\text{imporovment over \textsc{EP}})}$ for training on a $50/50$ data mixture of RedPajama and C\# code with longer $64$K context.}

  \end{center}
  
\end{table}
\twocolumn
\section{Ablations}

\subsection{\method{} Parameters}\label{sec:appendix_hypparam_splice}
There are two important design choices related to \method{}. First, how many related documents are retrieved in each step (the parameter $k$ in Algorithm \ref{alg:splice}). Second, how the documents are ordered. Table \ref{tab:ablationsplice} indicates that $k=1$ is the best choice for naturally ocurring long documents whereas Table \ref{tab:ablationsplice_k16} shows that in case of RAG-style input greater values of $k$ achieve better performance, though the differences are rather small. We found that changing the order of documents in training samples hardly matters for $270$M models. We use 'standard', as ordered by Algorithm \ref{alg:splice}, the reversed order, and random shuffling.

\begin{table*}
  
  \begin{center}
  \begin{tabular}{c c c | c  c c c c}
    \toprule
    & & \multicolumn{5}{c}{\textbf{Code}} \\
    \textbf{Method} & \textbf{Top-$k$} & \textbf{Order}  & C++ & Haskell  & Python & CUDA & All    \\
    \midrule
    \multirow{7}{*}{\methodvariant{\bmtf{}}} & \multirow{2}{*}{top 1} & standard        & \bf{2.38} & \bf{3.20} & \bf{2.82} & \bf{2.23} & \bf{2.93}   \\
    &  & reverse & \bf{2.38} & 3.21 & \bf{2.82} & \bf{2.23} & \bf{2.93}    \\
    \cmidrule{2-8}
    & \multirow{2}{*}{top 2} & standard & 2.39 & 3.23 & 2.84 & 2.24 & 2.95 \\
    &  & reverse & 2.39 & 3.24 & 2.84 & 2.24 & 2.95 \\
    \cmidrule{2-8}
    & \multirow{3}{*}{top 3} & standard        & 2.39 & 3.24 & 2.84 & 2.25 & 2.97    \\
    &  & reverse & 2.39 & 3.25 & 2.84 & 2.25 & 2.96   \\
    &  & shuffle & 2.40 & 3.24 & 2.84 & 2.25 & 2.96   \\

    \bottomrule
  \end{tabular}
  \end{center}
  \caption{\small
  Ablation of \method{} hyper-parameters. For each ablation, we have trained the same $270M$ parameter model using different data organization methods.
  Top-$k$ corresponds to the number of descendants chosen in the $\texttt{RETRIEVE}(d, k)$ step of the Algorithm \ref{alg:splice}.
  Reverse and shuffle correspond to the final order of examples $C$ returned by the Algorithm \ref{alg:splice} (reverse -- the order of documents in $C$
  is reversed, shuffle -- examples are shuffled.}
  \label{tab:ablationsplice}
\end{table*}

\begin{table*}
\centering
\begin{tabular}{c|c c c c c}
\hline
\textbf{Method}& & \multicolumn{3}{c|}{\textbf{Code}} & \textbf{Code \&} \\
 & \textbf{arXiv} & \textbf{C RAG-style} & \textbf{Python} & \textbf{Code} & \textbf{arXiv} \\
\hline
 \methodvariant{\bmtf{}} k=1 & \textbf{5.46} & 3.20 & \textbf{2.81} & \textbf{2.94} & \textbf{3.10} \\
 \methodvariant{\bmtf{}} k=16 & 5.49 &\textbf{ 3.18} & 2.85 & 2.98 & 3.14 \\
\hline
\end{tabular}
\caption{ Ablation of \method{} hyper-parameter $k$, we can observe that increasing $k$ can bring benefits on RAG-style input data (input created by retrieving related documents with \method{} $k=16$) at a cost of performance on natural long documents. Models were tuned on a 50/50 mixture of C prepared with \method{} and RedPajama}\label{tab:ablationsplice_k16}
\end{table*}

We also evaluated the influence of \texttt{BOS} and \texttt{EOS} tokens on the performance of trained models. As both \textsc{Repo} and \method{} methods concatenate documents to create training samples, they effectively slightly decrease the number of separating tokens compared to the \textsc{EP}. However, in Appendix \ref{sec:appendix_boseos} we included experiments showing that this has no impact on model performance.

\subsection{Importance of Separating Tokens and In-Domain Sampling}\label{sec:appendix_boseos}
We also evaluate the influence of \texttt{BOS} and \texttt{EOS} tokens on the performance. 
To be more precise, in all setups, training samples are separated by \texttt{BOS} and \texttt{EOS} tokens. 
As \method{} methods concatenate documents to create training samples, they effectively increase the average example length and decrease the number of separating tokens.
To check whether those methods do not simply benefit from the reduction in the number of \texttt{BOS} and \texttt{EOS} tokens and concatenation of documents from the same domain, we have trained a model on data prepared similarly as in \method{}, but instead of most matching documents $\texttt{RETRIEVE}(d, k)$ returned random documents from the C dataset (sampling without replacement). The results are shown in Table \ref{tab:ablationrandom}. We note that the difference between the \textsc{EP} and the random concatenation approach is small, and the random concatenation approach does not result in significant perplexity gains. When concatenating the documents, we concatenate them within the domain (C language in this case).
\begin{table*}

  \begin{center}
  \begin{tabular}{c c | c | c c c c | c}
    \toprule
    \textbf{Altered} & \textbf{Method} & \multicolumn{1}{c}{\textbf{arXiv}} & \multicolumn{2}{c}{\textbf{Code}} & \multicolumn{1}{c}{\textbf{Code \&}} \\
    \textbf{Data} &  &  & Python & All & \textbf{arXiv}  \\
    \midrule
    \multirow{2}{*}{C} & \textsc{WithinDomEP} & 5.554 $\pm$ 0.004 & \textbf{2.931} $\pm$ 0.003 & 3.076 $\pm$ 0.005 & 3.231 $\pm$ 0.005 \\
    & \textsc{EP} & \textbf{5.550} $\pm$ 0.002 & \textbf{2.931} $\pm$ 0.008 & \textbf{3.073} $\pm$ 0.006 & \textbf{3.228} $\pm$ 0.005 \\
    
    \bottomrule
  \end{tabular}
  \end{center}
  \caption{\small Perplexity evaluation of two methods of organizing the data. \textsc{EP} -- example packing, document equals training sample. \textsc{WithinDomEP} -- concatenate documents within the domain into examples of length bounded by $120$K characters. Training samples are then fed into the model and separated by \texttt{BOS} and \texttt{EOS} tokens.
  The difference is negligible, which suggests that the extension of example length in \textsc{WithinDomEP} does not help the model to utilize extended context. Experiments were performed using the $270$M parameter model. We performed three runs on different subsets of C to provide mean and standard deviation. The training data is a 50/50 mixture of RedPajama, organized in a standard way, and C data, organized using a method of choice.}
  \label{tab:ablationrandom}
\end{table*}

\onecolumn
\section{Key-Value Retrieval Task}\label{sec:appendix_kv_retriv_nelson}
Figure \ref{fig:lit300k7bcl} shows how training on \method{} organized data improves the performance on the key-value retrieval task proposed in \citep{liu2023lost}. This is a zero-shot task in which a model is prompted with a JSON dictionary and asked to retrieve a value corresponding to a specified key. The structure of the input is showcased below. 

\begin{scriptsize}

\begin{center}
\begin{verbatim*}
Extract the value corresponding to the specified key in the JSON object below.

JSON data:
{"4ef217b7-6bc0-48c6-af35-2765f1e730f3": "068192b7-16b1-40e0-8495-61c63f979d50",
 "cd6b8bdc-bc6c-4490-acb4-bc187a2dccba": "7364a26e-289f-4968-93d3-b273e882bdee",
 "7d057372-4ab8-4811-8110-658c3f19fff4": "3ad075c5-b567-4201-85a7-cb31a0c91540",
 "c62e192d-45e6-4646-bb88-1529c73256c9": "f0411644-1f6d-42a6-8af8-f06da66efc77",
 "06134e93-e158-490e-a66c-8e3b98e12735": "50a26a36-d832-450c-8d6e-a4cc3d0ec0ab",
 "3286f978-4270-4b54-8bfa-540d7e0772e6": "075cc716-1836-4f90-9be3-53e3d4ec6585",
 "4701aa05-c523-4b89-9700-64ab9c37c537": "49d86354-74c4-4256-9b3a-35e6e2b80d00",
 "c8895805-e574-4f13-9fe5-89da1d8c4748": "cc91af7f-8509-4bdc-bad7-2646af68e6d2"}
 "4701aa05-c523-4b89-9700-64ab9c37c537":

\end{verbatim*}
\end{center}

\end{scriptsize}

\begin{figure}[H]
  \centering
\begin{minipage}{0.4\textwidth}
\centering
     \includegraphics[scale=0.2]{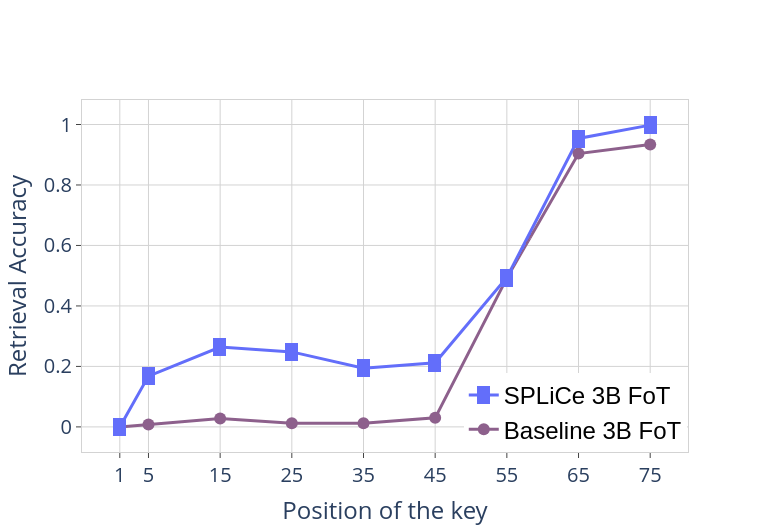}
      \subcaption{}
      \label{fig:lit75subfig3bfot}
\end{minipage}
\begin{minipage}{0.4\textwidth}
\centering
     \includegraphics[scale=0.2]{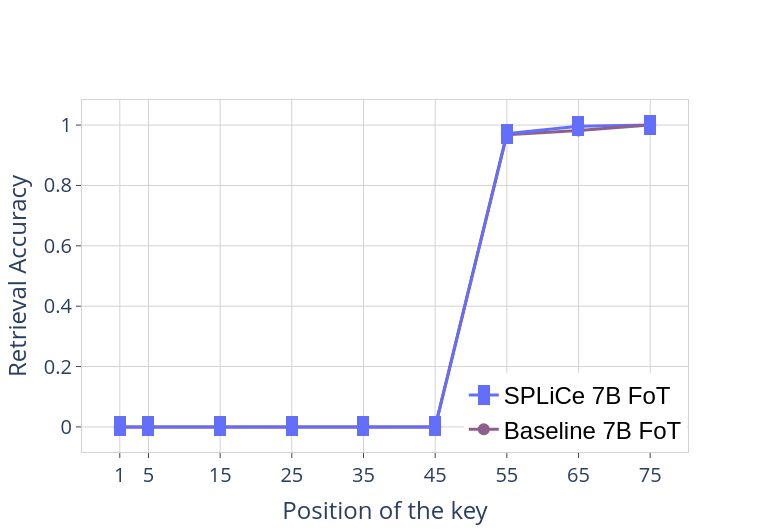}
      \subcaption{}
      \label{fig:lit75subfig7bfot}
\end{minipage}
\begin{minipage}{0.4\textwidth}
\centering
     \includegraphics[scale=0.2]{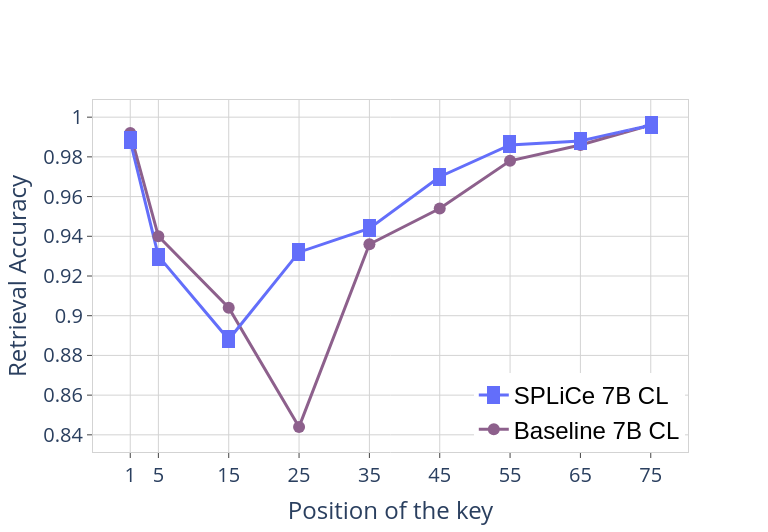}
      \subcaption{}
      \label{fig:lit75subfig7bcl}
\end{minipage}
\caption{Performance on a smaller version of key-value retrieval task from \citep{liu2023lost}. We note that FoT models (a), (b) generally struggle to retrieve tokens that are only visible to a subset of layers with extended context. For comparison, we show the results with a model that has extended context in all layers (c) using CodeLlama \citep{roziere2023code} method of context extension. Each position was evaluated using $500$ examples. }
\end{figure}

We noted that FoT-trained models struggle with this task. This is probably due to the fact that they extend context only in a couple of layers, and the key-value retrieval task requires looking up and extracting a long sequence of letters and digits. Because of that, we evaluate FoT models with shorter dictionaries consisting of $75$ key-value pairs (around $6$K tokens) and show the results in Figures \ref{fig:lit75subfig3bfot}, and \ref{fig:lit75subfig7bfot}. For comparison, we also evaluate the $7$B CL model with this context length and show the results in Figure \ref{fig:lit75subfig7bcl}.

\twocolumn

\section{Perplexity Improvements}\label{sec:appendix_3B_perplexity}
\begin{figure}[h]
  \centering
  \includegraphics[width=0.5\textwidth]{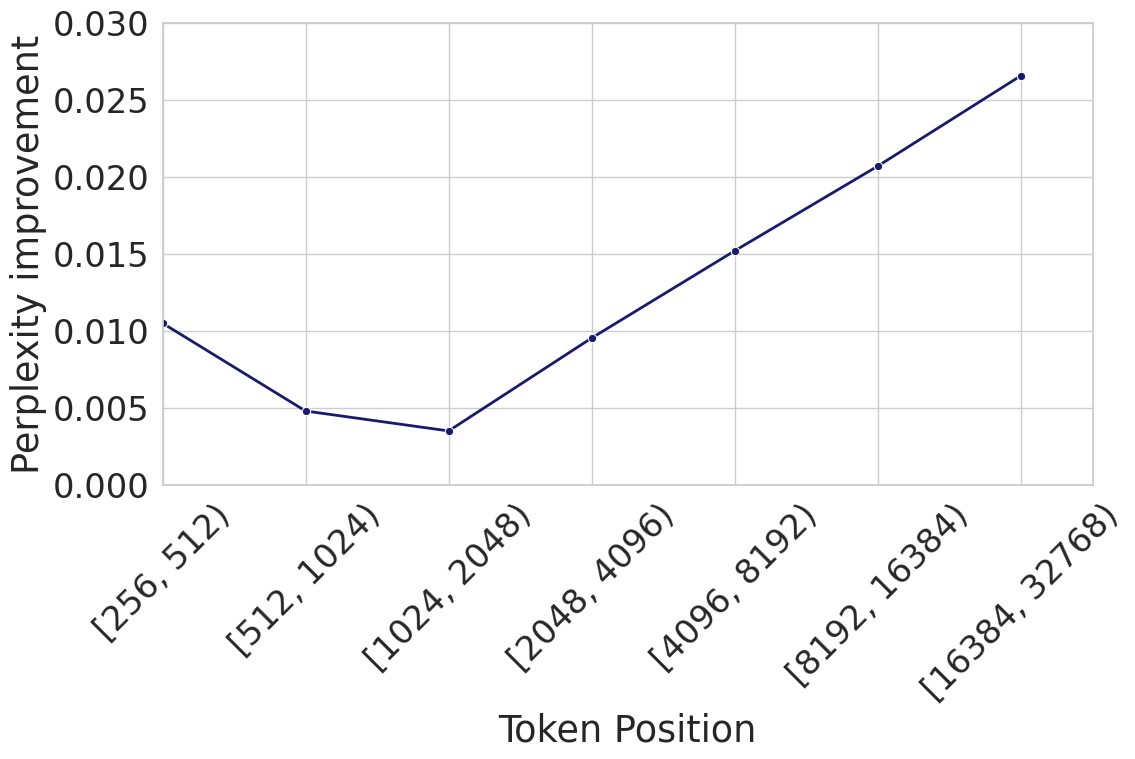}
  \caption{\small Perplexity improvement with \method{} against the \textsc{EP} of the final models (after $21$k training steps). We bucket tokens by their positions in the document and calculate the average. Each dot is the difference of the averages of the \method{} and \textsc{EP} models. We observe that \method{} has smaller perplexity, and the improvements tend to be larger for tokens further in the document.  }
  \label{fig:ppl_one}
\end{figure}

\begin{figure}[h]
  \centering
  \includegraphics[width=0.5\textwidth]{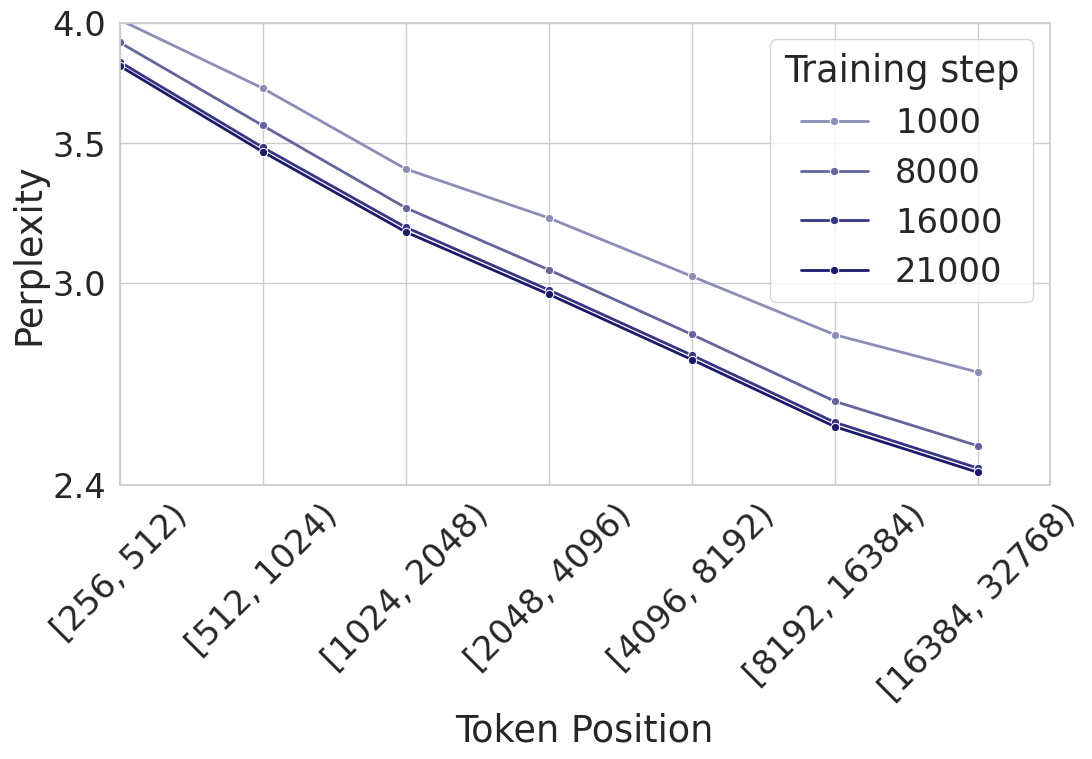}
  \caption{Evolution of the perplexity with \method{}, as the model is trained on more tokens. See \ref{fig:ppl_one} for the difference with the baseline. As expected, \method{} significantly improves perplexity for tokens whose positions are very distant in the sequence. Perplexity for more distant tokens improves more significantly compared to tokens in the beginning, early in the training. For more see Table \ref{tab:trecsteps3bfotsplice}.}
  \label{fig:ppl_both}
\end{figure}

{In Figure \ref{fig:ppl_one} we present perplexity improvements of $3$B FoT \methodvariant{\bmtf{}} over \textsc{EP}. Figure \ref{fig:ppl_both} shows the evolution of \method{} model perplexity during the training.} We follow \citep{anthropic2023modelcard} and bucket perplexity by token positions in buckets of length $2^i$ up to $32768$, and then average within the buckets. We average perplexity across arXiv, CUDA, Haskell, and CommonCrawl datasets.

\begin{table}

  \begin{center}
  \begin{tabular}{c | c  c c }
    \toprule
    \textbf{Steps} & \textbf{4K} & \textbf{8K} & \textbf{16K} \\
    \midrule
    TREC & {64.2 {\small $\pm$} 5.3}  & {72.9 {\small $\pm$} 4.1} & {72.9 {\small $\pm$} 4.1} \\
    
    \bottomrule
  \end{tabular}
  \end{center}
  \caption{\small Evaluation of the 3B \method{} model on  780-shot TREC using 10 seeds and different checkpoints.
  See Figure \ref{fig:ppl_both} for perlexity.}
  \label{tab:trecsteps3bfotsplice}
\end{table}

\section{Data Preparation}\label{sec:apendix_data_prep}
\subsection{Evaluation Data}\label{sec:apendix_eval_data}
We have taken a random subset of arXiv from Proof-pile.
For StarCoder data, we have downloaded up to $64$GB of each of the mentioned language subsets and performed a random 85/15 split for languages that we train on.

When evaluating the perplexity of the model, we skip documents that are shorter than the model context and truncate documents that are longer than that.
Table \ref{tab:eval_tokens} shows the number of tokens over which the perplexity was calculated.

\begin{table}[H]

\begin{center}
\begin{tabular}{llll}
\toprule
Eval/Method & 16K& 32K & 64K\\
\midrule
ArXiv & 16M & 16M & 16M \\
C & 16M & 16M & 16M \\
C++ & 16M & 16M & 16M \\
CUDA & 16M & 14M & 8M \\
C\# & 16M & 16M & 16M \\
Common Lisp & 16M & 16M & 16M \\
Dart & 16M & 16M & 16M \\
Emacs Lisp & 16M & 15M & 9M \\
Erlang & 16M & 16M & 12M \\
Fortran & 16M & 16M & 16M \\
Go & 16M & 16M & 16M \\
Groovy & 11M & 4M & 2M \\
Haskell & 16M & 16M & 15M \\
Java & 16M & 16M & 16M \\
Pascal & 16M & 16M & 16M \\
Python & 16M & 16M & 16M \\
\bottomrule
\end{tabular}
\end{center}
\caption{\small Number of evaluation tokens in each of the considered datasets. For each context length $c$, we consider only documents that have not less than $c$ tokens and extract the $c$ tokens prefix.}
\label{tab:eval_tokens}
\end{table}

\subsection{Train Data}
The StackExchange data was taken from the Proof-pile.
To prepare the code train data, we take the StarCoder train splits mentioned in Section \ref{sec:apendix_eval_data}, shuffle them, group the documents by the repository (documents from the same repository occur one after another), and split them into smaller packs. 
We also split repos larger than $25$MB and filter out files that are longer than $30$k characters. The reason behind repo splitting is to avoid the situation where one repository occupies a significant portion of the data pack. We have noticed repos containing as many as $40$K files and files as long as $11$M characters. The character filtering is consistent with our method as we aim to improve the performance in a scenario that lacks high-quality long-context data.
For C\# and Python, only one pack is used to organize the data. For C, we have performed a run on three packs and provided results and standard deviation in Table \ref{tab:cvariancecheck}. For large models, we run the methods on several packs and concatenate the results into a single dataset.
For natural language datasets, we extract a random subset of documents.

\section{{Faiss Parameters}}\label{sec:apendix_faiss}
{
Our experiments with \methodvariant{\contrieverms{}} utilize Faiss \citep{johnson2019billion}
for fast approximate inner-product search. To be more precise, we use the "IVF8192,Flat" index that we train on 262144 examples coming from the dataset.
}

\section{Detailed Accuracy Improvements}\label{sec:appendix_trec}
The performance of in-context learning depends much on the choice of the in-context examples. To study this in more detail we study the following random variable 
\[
  \Delta(c) = \text{ACC}_{\method}(c) - \text{ACC}_{\textsc{EP}}(c),
\]
where $\text{ACC}_{\method}(c), \text{ACC}_{\textsc{EP}}(c)$ are the accuracies of the model trained with \method{} and \textsc{EP} respectively on the random choice of in-context examples $c$. Below we report the histograms of $\delta(c)$. In Table \ref{tab:table_trec} and Table \ref{tab:table_db} we report mean $\Delta$ and $95\%$ confidence intervals as $\Delta${\scriptsize~[confidence interval]}.

\begin{figure}[H]
  \centering
\begin{minipage}{0.45\textwidth}
\centering
     \includegraphics[scale=0.22]{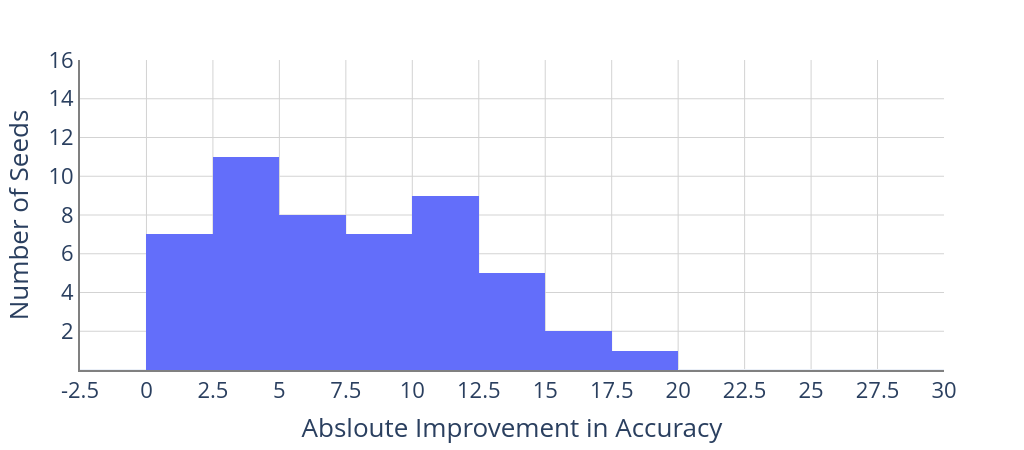}
      \subcaption{Examples: $190$, Context $4$K}
\end{minipage}
\begin{minipage}{0.45\textwidth}
\centering
     \includegraphics[scale=0.22]{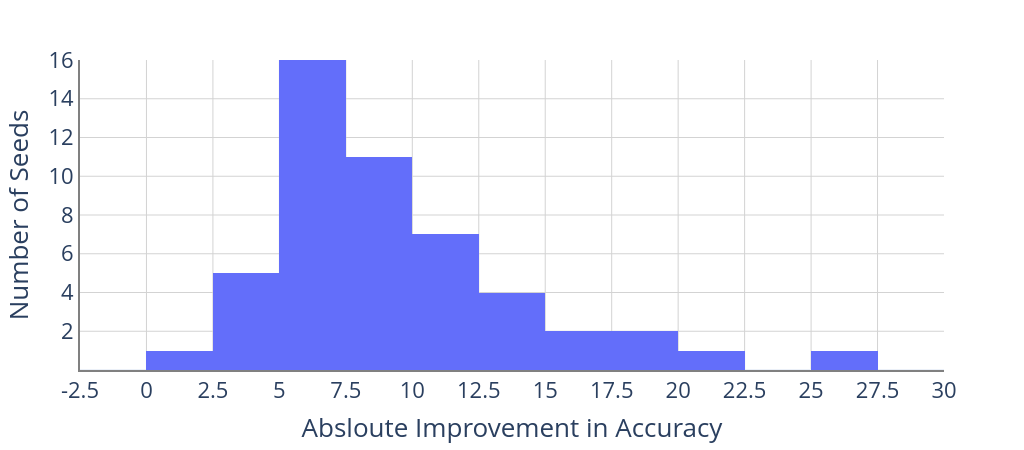}
      \subcaption{Examples: $380$, Context $8$K}
\end{minipage}

\begin{minipage}{0.45\textwidth}
\centering
     \includegraphics[scale=0.22]{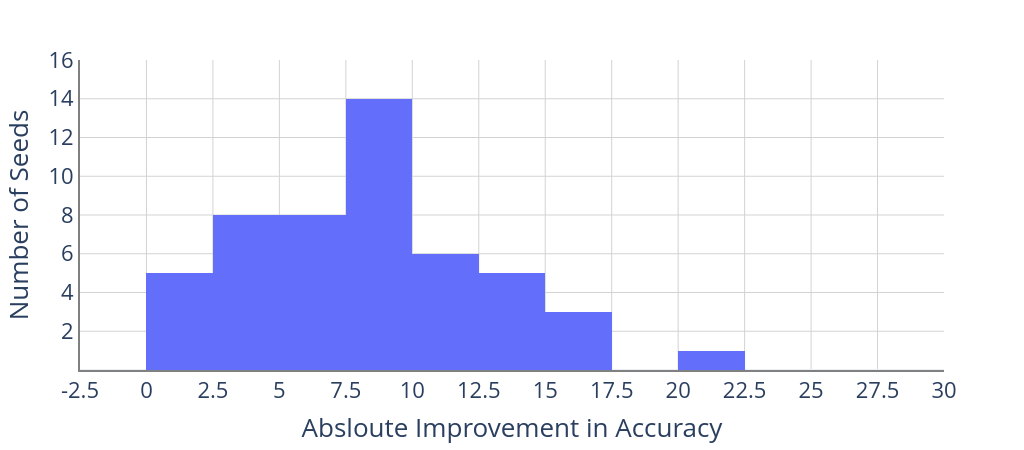}
      \subcaption{Examples: $780$, Context $16$K}
\end{minipage}
\begin{minipage}{0.45\textwidth}
\centering
     \includegraphics[scale=0.22]{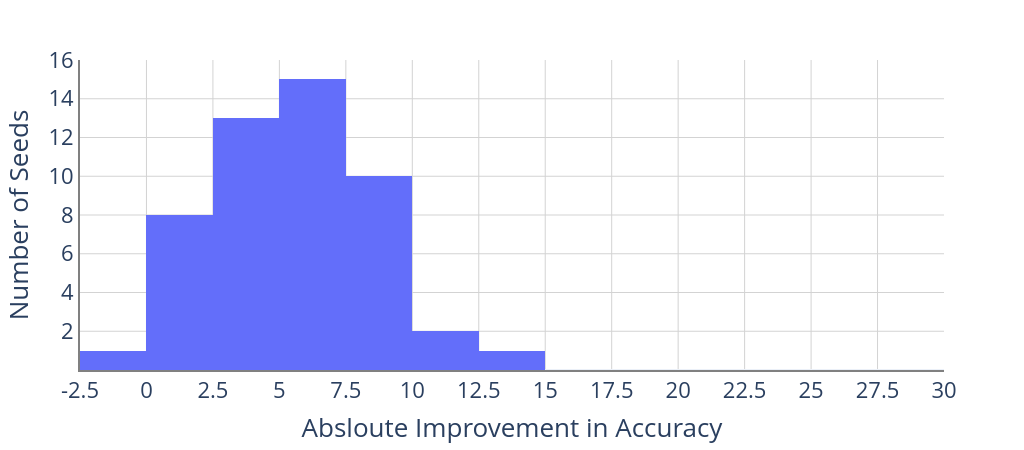}
      \subcaption{Examples: $1560$, Context $32$K}
\end{minipage}
  \caption{{Histograms of accuracy improvement of \methodvariant{\bmtf{}} over \textsc{EP} on TREC question classification task. The results are obtained by comparing the accuracy on the test set of TREC of the $3$B FoT model trained with \method{} to the model trained with default data preparation method (\textsc{EP})  across $50$ sets of in-context examples. Each set of in-context examples consists of elements randomly sampled (without replacement) from the training subset of TREC. Note that the model trained with \method{}  is almost always better than the \textsc{EP}.}}
  \label{fig:acc_3b_trec_many}
\end{figure}

Figure \ref{fig:acc_3b_trec_many} shows additional details about accuracy improvements on TREC when considering different numbers of in-context examples.

\begin{figure}[H]
  \centering
\begin{minipage}{0.45\textwidth}
\centering
     \includegraphics[scale=0.20]{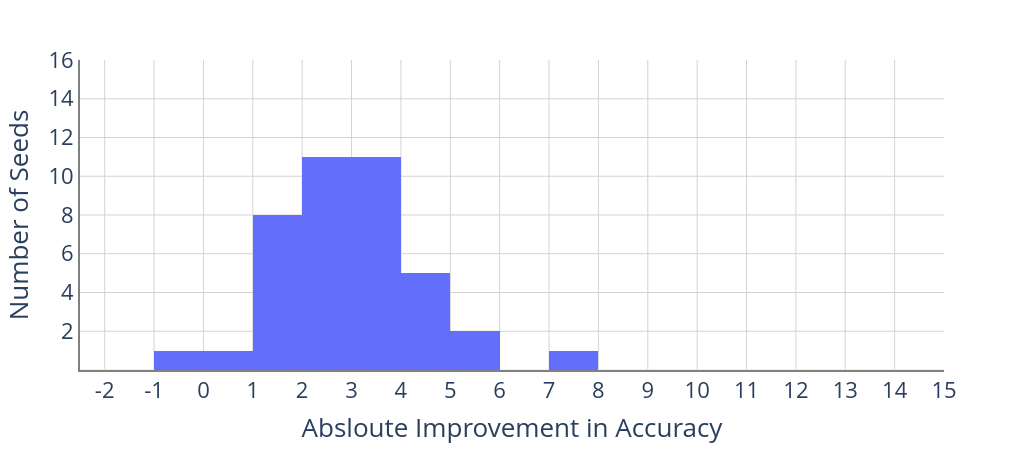}
      \subcaption{Examples: $190$, Context $16$K, Model: $3$B FoT}
\end{minipage}
\begin{minipage}{0.45\textwidth}
\centering
     \includegraphics[scale=0.20]{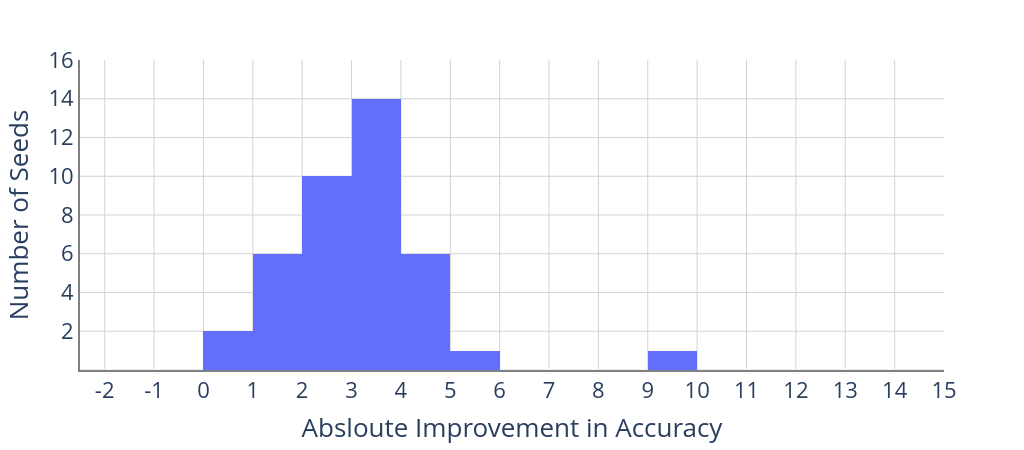}
      \subcaption{Examples: $380$, Context $32$K, Model: $3$B FoT}
\end{minipage}

\begin{minipage}{0.45\textwidth}
\centering
     \includegraphics[scale=0.20]{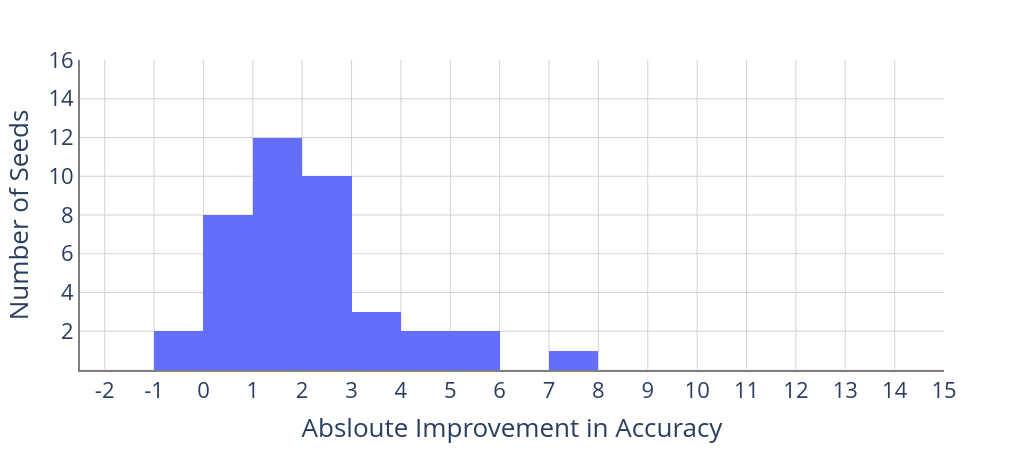}
      \subcaption{Examples: $190$, Context $16$K, Model: $7$B FoT}
\end{minipage}
\begin{minipage}{0.45\textwidth}
\centering
     \includegraphics[scale=0.20]{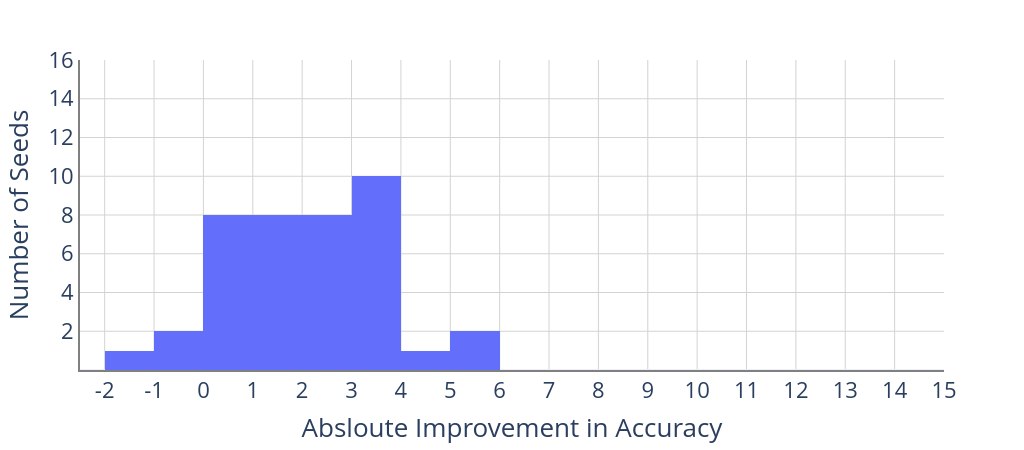}
      \subcaption{Examples: $380$, Context $32$K, Model: $7$B FoT}
\end{minipage}

\begin{minipage}{0.45\textwidth}
\centering
     \includegraphics[scale=0.20]{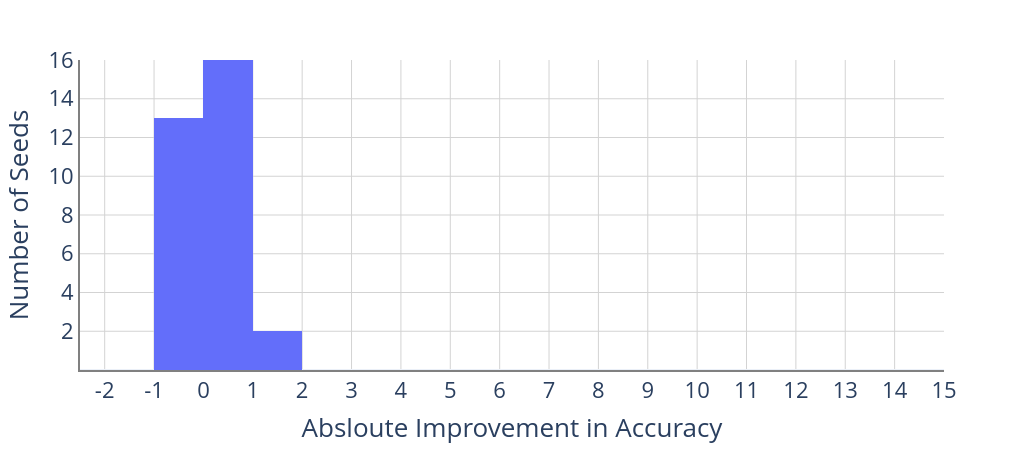}
      \subcaption{Examples: $190$, Context $16$K, Model: $7$B CL}
\end{minipage}
\begin{minipage}{0.45\textwidth}
\centering
     \includegraphics[scale=0.20]{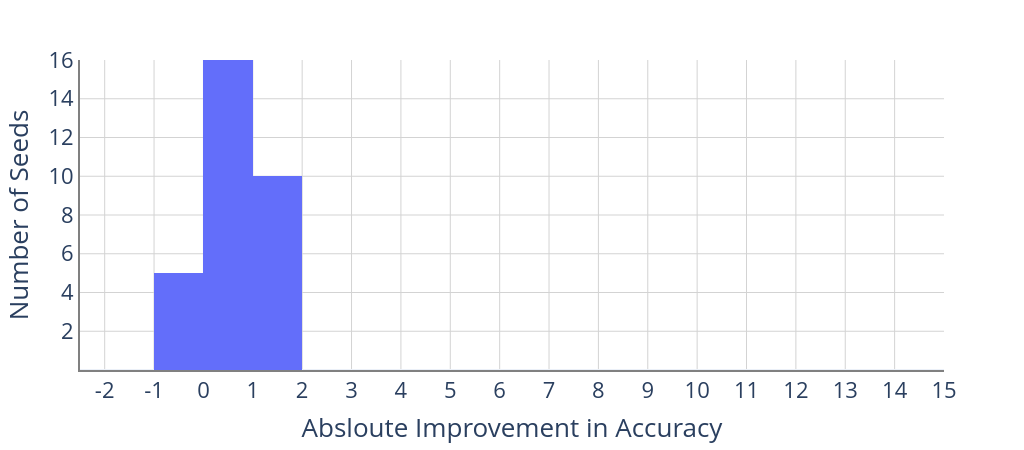}
      \subcaption{Examples: $380$, Context $32$K, Model: $7$B CL}
\end{minipage}
  \caption{Histograms of accuracy improvement of \methodvariant{\bmtf{}} over \textsc{EP} on DBPedia. We sample $40$ sets of in-context examples and, for each set of in-context examples, evaluate on a random $500$ element subset of the DBPedia test set.}
  \label{fig:acc_fot_dbpedia_many}
\end{figure}

\section{Results of $3$B Models on MMLU and GSM8K}\label{sec:apendix_table_shortctx_3B}
We present the missing results in Table \ref{tab:table_shortctx_3B}.
\begin{table}[h]
    \centering

    \begin{tabular}{l|ccc|c}
    \toprule
    \multicolumn{1}{c}{\textbf{Model}}& \multicolumn{3}{c|}{\textbf{MMLU}} & \textbf{GSM8K}\\
    & STEM & HUM & \multicolumn{1}{c|}{All} &  \\
    \midrule
    $3$B \textsc{ST CHKPT} & 24.4  & 25.6 & 25.9  & 4.3  \\
    $3$B$_{ \text{FoT}}$ \method{} & 25.8 & 26.7 &  26.5  & 2.5  \\
    $3$B$_{ \text{FoT}}$ \textsc{EP} & 25.2 & 26.2 & 25.7  & 3.4 \\
    \bottomrule
    \end{tabular}
    \caption{\small We evaluate our models on MMLU ($5$-shot), GSM8K ($8$-shot CoT). We provide an additional comparison with their starting checkpoint. In the main paper we note that the results of $3$B parameter model starting checkpoint are close to random and were moved this Appendix. See Table \ref{tab:table_shortctx} for results regarding larger models.}\label{tab:table_shortctx_3B}
\end{table}

\section{Ordering of Examples} \label{sec:large_models_shuffling}
We typically use the identity ordering as \texttt{Order} in Algorithm \ref{alg:splice} to merge documents into a single context, as we found that in most cases, it performs best. We found that random shuffling is slightly better in some cases. Specifically, this is the case of large $7$B CodeLlama models, see Table Table \ref{tab:table_trec_shuf_no_shuf}. We hypothesize that random ordering forces the model to make use of the full space of the RoPe positional encoding. Whereas the identity ordering, also used in \citep{shi2023incontext}, skews the model toward paying more attention to fragments of text that are not too far away. This suggests an interesting research direction on the intersection of data preparation and positional embeddings.

\begin{table}[H]

  \begin{center}
    \begin{tabular}{c c|cccc}
    \toprule
    \multicolumn{4}{c}{TREC}\\

    \midrule
    Model & \begin{tabular}{@{}c@{}}Context length \\ {\small (\# of examples})\end{tabular}  & \method{}-{\tiny no-shuf} & \method{}-{\tiny shuf} \\

    \midrule
    
    \multirow{2}{*}{$7$B CL} 
    & $32$K { (1560)} & 76.0 { $\pm$2.5} & 76.6 { $\pm$1.9}\\ 
    & $8$K { (380)} & 77.1 { $\pm$2.4} & 76.5 { $\pm$1.8}\\
    \bottomrule
    \end{tabular}
  \end{center}
  \caption{\small Average classification performance on TREC \citep{li-roth-2002-learning, hovy-etal-2001-toward}. We compare $7$B CL model trained on \method{} prepared data with different approaches to ordering the examples (different function \texttt{Order} in \ref{alg:splice}). \method{}-{\tiny shuf}  denotes the model trained on data that shuffled the documents randomly in context, for \method{}-{\tiny no-shuf} \texttt{Order} was the identity function. We use $\pm$ to denote the standard deviation. We decided to stick with the model trained with random shuffling as it has slightly better long-context performance and lower standard deviation.}
  \label{tab:table_trec_shuf_no_shuf}
\end{table}

\section{HotPotQA and Qasper Results}\label{sec:appendix_qasper_hpqa}
We evaluate our models on HotPotQA \citep{hotpotqa} and Qasper \citep{dasigi2021dataset} from SCROLLS \citep{shaham2022scrolls} on both long (see Table \ref{tab:table_qasper_hpqa}) and short context (see Table \ref{tab:table_qasper_short} and Table \ref{tab:table_hpqa_short})

 \begin{table*}
   
   \begin{center}
     \begin{tabular}{c |ccc|cccc}
     \toprule
     \textbf{Task} & \multicolumn{3}{c|}{\textbf{QASPER}} & \multicolumn{4}{c}{\textbf{HotPotQA}}\\
     \midrule
    \small Model & Context &  \small \textsc{EP} & \small \method{} & Context & \small \textsc{EP} & \small \method{} & $\Delta ${\scriptsize~[conf interv]} \\
 
     \midrule
     $3$B$_{ \text{FoT}}$ & \multirow{4}{*}{32K} & 22.1 & \textbf{22.8} & \multirow{4}{*}{20K} & \textbf{29.7}  & 29.6 & -0.1~\scriptsize[${-0.3,0.1}]$ \\
 
     $7$B$_{ \text{FoT}}$ &  & 22.8 & \textbf{23.1} &  & 26.0  & \textbf{27.3} & 1.3~\scriptsize[${1.2,1.5}]$ \\
 
     $7$B$_{ \text{CL}}$  &  & 29.0 & \textbf{29.7} &  & 31.0  & \textbf{31.2} & 0.2~\scriptsize[${0.1,0.6}]$ \\
     $13$B$_{ \text{FoT}}$ &  & \textbf{32.0}  & \textbf{32.0}  &  & 36.5   & \textbf{36.7}   & 0.2~\scriptsize[${0.1,0.4}]$ \\
     \bottomrule
     \end{tabular}
   \end{center}
   \caption{\small  We measure question answering over long input using Qasper \citep{dasigi2021dataset} (2-shot setting)
   and HotPotQA \citep{hotpotqa} (10-shot setting). For Qasper, we use the implementation from Language Model Evaluation Harness \citep{eval-harness}. For HotPotQA, we average results across $7$ sets of in-context examples for $3$B and $7$B models and $2$ for $13$B ones, with $\Delta${\scriptsize[confidence interval]} denoting mean improvement and its $95\%$ bootstrap confidence intervals. We use F1 score for evaluation. Note that in the $3$B model case, despite using \method{} for code data only, we still have improvements in non-code tasks.} 
   \label{tab:table_qasper_hpqa}
 \end{table*}

 \begin{table}
    \centering
    \begin{tabular}{l c c c}
        \toprule
        \multicolumn{4}{c}{QASPER}\\
        \midrule
        \textbf{Model} & \textbf{Context} & \textbf{Method} & \textbf{Result} \\
        \midrule
        \multirow{5}{*}{3B FoT} 
        & 32K & SPLiCe & \textbf{22.8} \\
        & 2K & SPLiCe & 18.4 \\
        & 2K & \textsc{EP} & 18.4 \\
        & 32K & ST CHKPT & 6.4 \\
        & 2K & ST CHKPT & 18.4 \\
        \midrule
        \multirow{4}{*}{7B FoT}
        & 32K & SPLiCe & \textbf{23.1} \\
        & 2K & SPLiCe & 18.0 \\
        & 2K & \textsc{EP} & 18.2 \\
        & 2K & ST CHKPT & 18.3 \\
        \midrule
        \multirow{4}{*}{7B CL}
        & 32K & SPLiCe & \textbf{29.7} \\
        & 2K & SPLiCe & 18.6 \\
        & 2K & \textsc{EP} & 18.4 \\
        & 2K & ST CHKPT & 18.3 \\
        \midrule
        \multirow{4}{*}{13B FoT}
        & 32K & SPLiCe & \textbf{32.0} \\
        & 2K & SPLiCe & 18.9 \\
        & 2K & \textsc{EP} & 19.1 \\
        & 2K & ST CHKPT & 19.9 \\
        \bottomrule
    \end{tabular}
    \caption{Comparison with starting checkpoint on Qasper. We note that the OpenLLaMA models were trained with context length 2K.}
    \label{tab:table_qasper_short}
\end{table}

\begin{table}
    \centering
    \begin{tabular}{l c c c}
        \toprule
                \multicolumn{4}{c}{HotPotQA}\\
        \midrule
        \textbf{Size} & \textbf{Context} & \textbf{Method} & \textbf{Result} \\
        \midrule
        \multirow{4}{*}{3B FoT} 
        & 32K & SPLiCe & \textbf{29.6} \\
        & 2K & SPLiCe & 26.4 \\
        & 2K & \textsc{EP} & 27.2 \\
        & 2K & ST CHKPT & 27.3 \\
        \midrule
        \multirow{4}{*}{7B FoT}
        & 32K & SPLiCe & 27.3 \\
        & 2K & SPLiCe & 21.0 \\
        & 2K & \textsc{EP} & 19.7 \\
        & 2K & ST CHKPT & \textbf{27.7} \\
        \midrule
        \multirow{4}{*}{7B CL}
        & 32K & SPLiCe & \textbf{31.2} \\
        & 2K & SPLiCe & 24.2 \\
        & 2K & \textsc{EP} & 24.3 \\
        & 2K & ST CHKPT & 27.7 \\
        \midrule
        \multirow{4}{*}{13B FoT}
        & 2K & SPLiCe & \textbf{36.7} \\
        & 2K & SPLiCe & 27.6 \\
        & 2K & \textsc{EP} & 25.6 \\
        & 2K & ST CHKPT & 32.9 \\
        \bottomrule
    \end{tabular}
    \caption{Comparison with starting checkpoint on HotPotQA (0-shot - 2k context and 10-shot 20k context). We note that the OpenLLaMA models were trained with context length 2K.}
    \label{tab:table_hpqa_short}
\end{table}

 \begin{table*}
   
   \begin{center}
     \begin{tabular}{c c | c c}
     \toprule
     \multicolumn{2}{c|}{\textbf{Task}} & \multicolumn{1}{c}{\textbf{Landmark Attention}} & \multicolumn{1}{c}{\textbf{RULER}}\\
     \midrule
       Model & Method & Passkey Retrieval & Variable Tracking  \\
 
     \midrule
     \multirow{2}{*}{$7$B$_{ \text{CL}}$ } & \method{} & 69.9 & 51.8 \\
      & \textsc{EP} & 68.6 & 29.8 \\
    
     \bottomrule
     \end{tabular}
   \end{center}
   \caption{\small  We additionally measure our $7$B$_{ \text{CL}}$ on passkey retrieval from Landmark Attention \citep{mohtashami2023landmark} and Variable Tracking from RULER \citep{hsieh2024rulerwhatsrealcontext}. We note that this model was trained on fewer tokens than the $3$B parameter FoT one and that none of our models were instruction-tuned. The results for  Passkey Retrieval were averaged over 1000 samples using 32K context length, for Variable Tracking we have utilized a shorter context length.} 
   \label{tab:table_qasper_hpqa}
 \end{table*}

\section{HumanEval}\label{sec:apendix_human_eval}
We perform an additional evaluation using HumanEval \citep{chen2021evaluating} and present the results in Table \ref{tab:table_shortctx2}.
We note that similarly to GSM8K results from Table \ref{tab:table_shortctx} \method{} improves the short context performance of the larger model.
\begin{table}
  \centering

  \begin{tabular}{l | l | c}
  \toprule
          \multicolumn{3}{c}{HumanEval}\\
        \midrule
  \textbf{Model}& Method & Pass@1\\
  \midrule
  $7$B & \textsc{Starting Checkpoint} & 12.9 \\
  \midrule
  \multirow{2}{*}{$7$B$_{ \text{FoT}}$} & \method{} & 10.0 \\
   & \textsc{EP} & 9.1 \\

   \midrule

  \multirow{2}{*}{$7$B$_{ \text{CL}}$} & \method{} & 8.5 \\
  & \textsc{EP} & 8.5 \\
    \midrule
  $13$B & \textsc{Starting Checkpoint} & 23.8 \\
  \midrule
  \multirow{2}{*}{$13$B$_{ \text{FoT}}$} & \method{} & 27.4 \\
   & \textsc{EP} & 26.8 \\

  \bottomrule
  \end{tabular}
  \caption{We additionally evaluate our $7$B and $13$B parameter models on HumanEval \citep{chen2021evaluating}. For $7$B models, all methods experience a decrease in performance, but \method{} is better at maintaining short context performance. For $13$B, both \method{} and \textsc{EP} increase the performance. We note that in our evaluation we do not use special tokens designed for CodeLlama $13$B code completion. Instead, evaluate all models in the same way. Because of the different evaluation pipeline our results for \textsc{Starting Checkpoint} are not the same as the ones presented in \citep{roziere2023code}.}
  \label{tab:table_shortctx2}
  \end{table}

  \section{$270$M Models Training, Fine-Tuning and Evaluation Details} \label{sec:appendix_training_eval_proto}
  \looseness=-1 \paragraph*{Training protocol} Initially, we train with the $2$K context length on $6.3$B tokens from RedPajama \citep{together2023redpajama}. Subsequently, we fine-tune using $1$B tokens with the context extended to $32$K on a mixture of the original RedPajama data \citep{together2023redpajama} and long context data created using \method{}/\textsc{EP}. The amount of pre-training tokens is based on scaling laws from \citep{hoffmann2022training} and constants for GPT-like models in \citep{nanoGPT}.  
  
  \looseness=-1 \paragraph*{Fine-tuning protocol} We employ the Focused Transformer (FoT) \citep{tworkowski2023focused} for context extension method (unless stated otherwise). This approach is motivated by practical factors, viz. training with short context length expedites the process, while context scaling can be achieved by finetuning on a relatively small amount of tokens, as demonstrated by \citep{chen2023extending, tworkowski2023focused}. Loosely inspired by \citep{ouyang2022training,roziere2023code}, in the latter phase, long context data (i.e., prepared with \method{}) constitutes half of the mixture. We also check how \method{} works with other context extension methods in Appendix \ref{sec:med_diff_ctx}.
  
  \paragraph*{Evaluation} We measure perplexity on held-out portions of the arXiv \citep{azerbayev2022proofnet} and StarCoder \citep{li2023starcoder} datasets employing a context length of $32$K. The selection of these datasets is motivated by the fact that they can benefit from long-context information as demonstrated in \citep{chen2023extending, li2023starcoder}. For example, functions are often re-utilized, and similar terms are employed across papers. We exclude documents with fewer than $32$K tokens and truncate those exceeding this length. In Appendix \ref{sec:short_context_270} we evaluate our models using short context data confirming no performance degradation with respect to the base model. For information regarding the training and evaluation data, see Appendix \ref{sec:apendix_data_prep}.

\onecolumn
\section{Needle In A Haystack} \label{sec:appendix_niths}
We utilize the following prompt for evaluating $7$B parameter CL models on Needle In A Haystack \citep{niths}.
And the default needle that was used by the authors of the benchmark to evaluate models such as Anthropic’s Claude.
As a context document, we utilized the ``PaulGrahamEssays'' option from the Needle In A Haystack evaluator. We use recall of keywords from the ground truth answer as the evaluation metric.

\begin{scriptsize}

\begin{center}
\begin{verbatim*}
You are given the following Question and Context. Answer the Question using the information hidden inside the Context.

Question: {retrieval_question}
Context: {context}

Question: {retrieval_question}
Answer:

\end{verbatim*}
\end{center}

\end{scriptsize}

\section{TREC 2K Evaluation}\label{sec:appedix_trec_2_k}
\begin{table}[h]

    \begin{center}
      \begin{tabular}{c | cccc }
        \toprule
         \multicolumn{5}{c}{\textbf{TREC Context 2K (90-shot)}} \\
        \midrule
        
         Model &  \textsc{ST CHKP}  &  \textsc{EP} &  \method{} & $\Delta ${\scriptsize~[conf interv]} \\
        \midrule

        \multirow{1}{*}{$3$B$_{ \text{FoT}}$} 
        & 58.4 & 61.4 & \textbf{66.4} & 8.0~\scriptsize[${5.7, 10.0}]$ \\

        \multirow{1}{*}{$7$B$_{ \text{FoT}}$} 
        & 59.6 & 62.9 & \textbf{63.7} & 4.1~\scriptsize[${0.1, 6.6}]$ \\

        \multirow{1}{*}{$7$B$_{ \text{CL}}$} 
        & \textbf{59.6} & 59.1 & 59.1 & -0.5~\scriptsize[${-4.1, 1.3}]$ \\

        \multirow{1}{*}{$13$B$_{ \text{FoT}}$} 
        & 78.7 & 77.6 & \textbf{82.0} & 3.3~\scriptsize[${1.9, 4.7}]$ \\

        \bottomrule
        \end{tabular}
    \end{center}
    \caption{ We assess the short-context (2K) performance on TREC \citep{li-roth-2002-learning, hovy-etal-2001-toward}. We average across $10$ sets of in-context examples and use $\Delta${\scriptsize~[conf interv]} to denote the mean improvement of \method{} over the starting checkpoint and its $95\%$ bootstrap confidence intervals.}\label{tab:table_trec_db_short}
  \end{table}%

\end{document}